\documentclass[journal]{IEEEtran}
\usepackage{cite}

\usepackage{amsfonts}
\usepackage{array}
\usepackage{url}
\usepackage{amsmath}
\usepackage{amssymb}
\usepackage{color}
\usepackage{multirow}
\usepackage{setspace}
\usepackage{algorithm}
\usepackage{algorithmic}
\usepackage{lineno}
\usepackage{subcaption}

\ifCLASSINFOpdf
   \usepackage[pdftex]{graphicx}
   \graphicspath{{../pdf/}{../jpeg/}}
   \DeclareGraphicsExtensions{.pdf,.jpeg,.png}
\else
   \usepackage[dvips]{graphicx}
   \graphicspath{{../eps/}}
   \DeclareGraphicsExtensions{.eps}
\fi

\ifCLASSOPTIONcompsoc
  \usepackage[caption=false,font=normalsize,labelfont=sf,textfont=sf]{subfig}
\else
  \usepackage[caption=false,font=footnotesize]{subfig}
\fi

\begin{document}

\title{Dual Sparse Aggregation Transformer for Multispectral Object Detection}

\author{Wencong~Wu,~Xiuwei~Zhang,~Hanlin~Yin,~Hongxi~Zhang,~Yanning~Zhang,~\IEEEmembership{Fellow,~IEEE}
\thanks{This research is supported in part by the National Natural Science Foundation of China (62476220), National Key Research and Development Program of China (2023YFC3209305, 2023YFC3209304), Guangdong Basic and Applied Basic Research Foundation (2024A1515030186), and Natural Science Basic Research Program of Shaanxi (2024JC-DXWT-07, 2024JC-YBQN-0719). (Corresponding authors: Xiuwei Zhang; Yanning Zhang.)}
\thanks{Wencong Wu is with the School of Computer Science, Northwestern Polytechnical University, Xi'an 710072, China (Email: wencongwu@mail.nwpu.edu.cn).}
\thanks{Hongxi Zhang is with the School of Cybersecurity, Northwestern Polytechnical University, Xi'an 710072, China (Email: hongxi0209@mail.nwpu.edu.cn).}
\thanks{Xiuwei Zhang, Hanlin Yin and Yanning Zhang are with the School of Computer Science, the Shaanxi Provincial Key Laboratory of Speech and Image Information Processing and the National Engineering Laboratory for Integrated Aerospace-Ground-Ocean Big Data Application Technology, Northwestern Polytechnical University, Xi'an 710072, China (Email: xwzhang@nwpu.edu.cn; iverlon1987@nwpu.edu.cn; ynzhang@nwpu.edu.cn).}
}

\markboth{Manuscript submitted to IEEE Transactions on Circuits and Systems for Video Technology}%
{Shell \MakeLowercase{\textit{ et al.}}: Bare Demo of IEEEtran.cls for IEEE Journals Journal of \LaTeX\ Class Files,~Vol.~14, No.~8, August~2015}
\maketitle

\begin{abstract}
Transformer-based approaches have obtained excellent performance in multispectral object detection tasks due to their ability to model long-range dependencies and capture complementary information. However, previous transformer-based multispectral detection methods tend to use all available tokens for similarity calculation, which results in redundant information interaction from irrelevant areas, leading to degraded detection performance. To overcome this challenge, we propose a novel Dual Sparse Aggregation Transformer (DSAFormer) for multispectral object detection, which consists of a Dual Sparse Transformer (DSFormer) and a Learnable Addition Fusion Block (LAFB). Specifically, the DSFormer is designed to exploit and boost cross-modal complementary information, thereby improving detection performance. It incorporates three key components: A Spatial Sparse Multi-Head Cross-Attention (SSMHCA) mechanism selectively captures cross-modal relationships at the spatial level by reserving only the high query-key similarity scores, eliminating irrelevant interactions. A Channel Sparse Multi-Head Cross-Attention (CSMHCA) mechanism performs similar sparse calculations at the channel level to enhance feature representation and filter out low matching query-key. A Multi-Scale Feature Refinement Layer (MSFRL) is developed to aggregate hierarchical features and suppress redundant information. To effectively fuse multimodal features, the LAFB is introduced to aggregate intramodal and intermodal feature information by feature reweighting. Extensive experimental results have demonstrated that our proposed DSAFormer achieves better detection performance against state-of-the-art methods on four public datasets, including the MFAD, FLIR, M$^3$FD, and LLVIP. The source code of our DSAFormer will be released at https://github.com/WenCongWu/DSAFormer.
\end{abstract}


\IEEEpeerreviewmaketitle

\section{INTRODUCTION}
\IEEEPARstart{O}{bject} detection is an indispensable task in the field of computer vision, which aims to understand the categories and locations of objects in scenes. Most existing work \cite{Ren17, YOLOv5, YOLOv8, Wang24} performs detection tasks in visible light (VIS) datasets, but fails in conditions such as darkness, rain, and haze. Since infrared (IR) images are formed by using the thermal radiation emitted by objects, they are insensitive to the above environments and can successfully depict the edges and shapes of objects in real scenes. It should be noted that the lack of temperature difference between the objects and the surrounding environments can make it difficult to distinguish foreground and background areas in infrared imaging, but VIS images are able to clearly describe the foreground and background regions in such scenes. From the above analysis, it can be found that there is obvious complementarity between VIS and IR images, whose effective fusion can offer significant discriminative information for multispectral object detection and achieve the round-the-clock and all-scenario applications \cite{Hu25, Qian25, HuH25}.

With the development of deep learning, many convolutional neural networks (CNNs) based multispectral object detection approaches have been developed. For example, Zhang et al. \cite{Zhang21} presented a guided attentive feature fusion (GAFF) for multispectral object detection, where the intra-modality and inter-modality attention modules were used to enhance multimodal features and select high-quality VIS or IR features, respectively. Wu et al. \cite{Wu23} designed an adaptive multimodal feature fusion and cross-modal vehicle index (AFFCM) framework for VIS-IR object detection, where a multimodal adaptive feature fusion (MAFF) module was proposed to promote useful cross-modal feature fusion by utilizing the softpooling channel attention to compute multimodal feature weights of VIS and IR modalities and choose important features. Wang et al. \cite{WangW24} developed a cross-modal aerial remote sensing image object detection network to extract key features, in which the illumination perception and uncertainty-aware modules were designed to capture changes in light intensity and mitigate the impact of multimodal differences, respectively.

\begin{figure}[htbp]
	\begin{center}
		\includegraphics[width=0.495\textwidth]{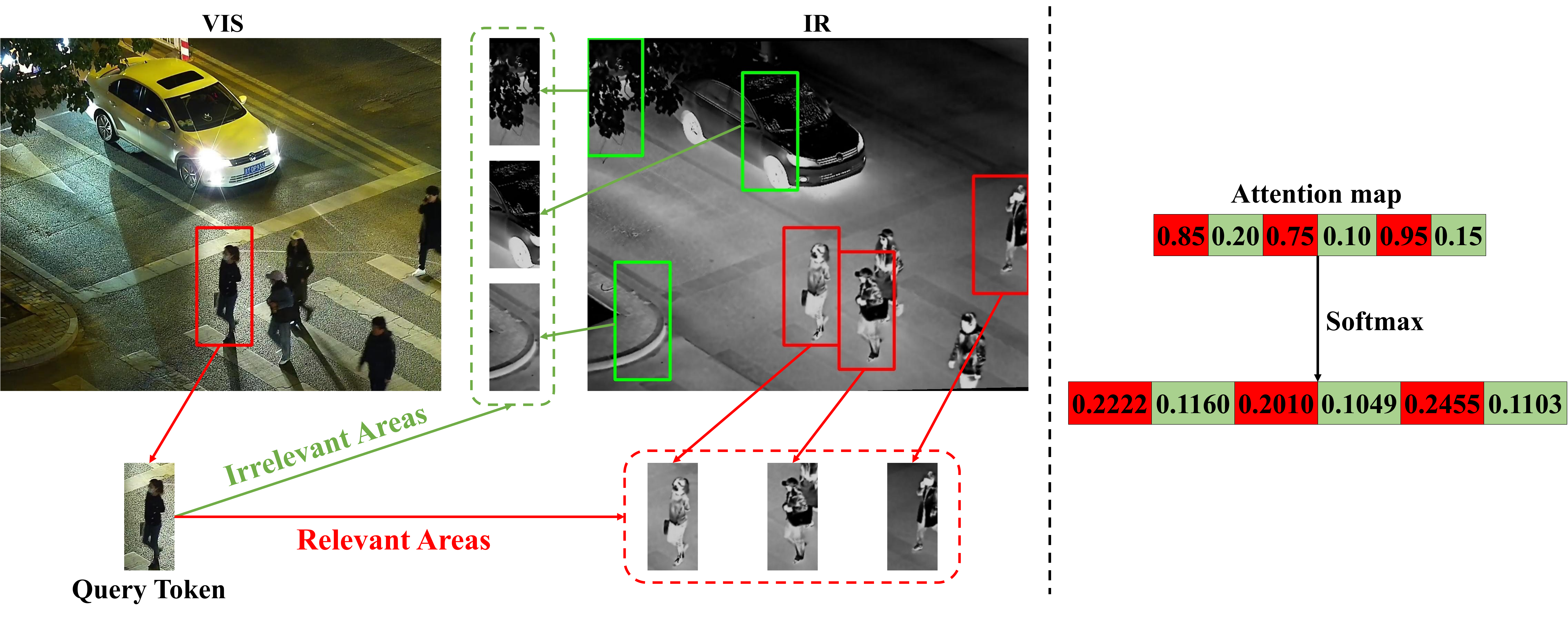}
		\caption{\textbf{Left image:} existing transformer-based multispectral detection methods perform token interactions in relevant and irrelevant areas between VIS and IR modalities, resulting in redundant information for similarity calculation. \textbf{Right image:} the information exchange between irrelevant regions across modalities generates low weights in the attention map, and the Softmax function will increase the value of these weights.}
        \label{fig0:motivation}
	\end{center}
\end{figure}

Although the above methods have achieved great detection results, they mainly focus on local information fusion and cannot establish long-distance dependencies, resulting in insufficient fusion of multimodal features and thus affecting detection performance. To this end, many transformer-based multispectral detection methods have been proposed. For instance, Fang et al. \cite{Fang21} constructed a cross-modality fusion transformer (CFT) to integrate different modalities for multispectral object detection, which can exploit long-range relationships and merge non-local information. Lee et al. \cite{Lee24} devised a multispectral object detection framework, namely CrossFormer, where a cross-guided attention mechanism (CGAM) was presented to calculate the attention maps by cross-modal feature interaction. Hu et al. \cite{HuH25} developed an edge-guided illumination-aware interactive learning-based detector (EI$^2$Det) for VIS-IR object detection, where the multi-head cross-attention mechanisms were applied to strengthen intra-modality and inter-modality feature representations, and the illumination-aware weighting module and edge-guided fusion module were respectively built to predict light intensity of VIS modality and employ important edge information to facilitate the detection performance. However, these transformer-based approaches utilize all tokens to generate similarity scores of query-key from different modalities, introducing irrelevant tokens into the calculation of attention maps, which results in redundant information interactions and thus reduces detection performance. As shown in the left image of Fig. \ref{fig0:motivation}, there are relevant and irrelevant areas between VIS and IR images. Token interactions across modalities in irrelevant regions can produce low similarity scores in the attention map, which are normalized by the Softmax function to increase the value and proportion of these scores, as displayed in the right image of Fig. \ref{fig0:motivation}. This will cause the detection model to distract from the object location, resulting in degraded detection performance.

To solve this issue, we propose a novel Dual Sparse Aggregation Transformer (DSAFormer) for multispectral object detection, where a Dual Sparse Transformer (DSFormer) is devised to capture cross-modal complementary information in the spatial and channel levels. In the DSFormer, a Spatial Sparse Multi-Head Cross Attention (SSMHCA) and a Channel Sparse Multi-Head Cross Attention (CSMHCA) are applied to exploit latent relationships in relevant query-key tokens based on the top-k selection mechanism in VIS and IR modalities, respectively. To further eliminate redundant information and extract multi-scale properties, we designed a Multi-Scale Feature Refinement Layer (MSFRL) to purify feature information and establish multi-scale relations. Moreover, we introduce a Learnable Addition Fusion Block (LAFB) to aggregate intramodal feature information and ensure the full integration of cross-modal feature information, which can improve the detection performance.

In brief, the main contributions of this work include the following three aspects:

(1) A novel Dual Sparse Aggregation Transformer (DSAFormer) is designed to successively exploit effective complementary features in the spatial and channel levels from different modalities and better integrate these feature information by a Learnable Addition Fusion Block (LAFB), which is highly competitive with other state-of-the-art approaches on four public datasets, including the MFAD, FLIR, M$^3$FD, and LLVIP.

(2) A Spatial Sparse Multi-Head Cross Attention (SSMHCA) and a Channel Sparse Multi-Head Cross Attention (CSMHCA) are designed to mitigate feature interaction of unimportant tokens by utilizing the mixed top-k selection mechanism to obtain relevant tokens from different modalities to boost feature aggregation.

(3) A new Multi-Scale Feature Refinement Layer (MSFRL) is proposed to capture the multi-scale relations and eliminate feature redundancy, which can enhance local information representation and improve the detection performance.

The remainder of this paper is arranged as follows. Section \ref{Sec_Related_work} reviews previous work related to our DSAFormer for multispectral object detection. In Section \ref{Sec_Methodology}, we introduce our proposed DSAFormer in detail. Next, Section \ref{Sec_Experiments} lists public datasets, model settings, evaluation metrics, and experimental results. Finally, we make a conclusion about our work in Section \ref{Sec_Conclusion}.

\section{RELATED WORK}\label{Sec_Related_work}
\subsection{CNN for Multispectral Object Detection}
Monospectral object detection methods \cite{YOLOv5, YOLOv8, Wang24, Girshick14, Girshick15, He20} tend to produce false and missed detection results in real complex scenes, such as dark night, thick fog, and other environments. Multispectral object detection approaches \cite{Zhou20, Zhang19, Cao23, Zhang23} can capture complementary cross-modal features, and effectively fusing these features is beneficial to obtain more accurate and robust detection performance.

For example, Liu et al. \cite{LiuL22} proposed a deep cross-modal representation learning-based pedestrian detection network, where a cross-modal feature learning module and a modality attention-based fusion strategy were used for extracting common and specific features from VIS-IR modalities and merging the detection results of the dual branch, respectively. Yan et al. \cite{Yan23} developed a cross-modality complementary information fusion network for multispectral pedestrian detection, which models the long-range dependencies and exploits the inter-spatial relationship between VIS and IR modalities. Fang et al. \cite{Fang22} designed a cross-modality attentive feature fusion for multispectral object detection, where a differential enhancive module and a common selective module were applied to extract modality-specific and modality-shared features respectively to promote cross-modal feature fusion and decline redundant information. Zhang et al. \cite{ZhangY23} presented an object detection network, named TINet, to integrate complementary features from VIS-IR modalities for multispectral object detection, in which an illumination-guided feature weighting module was adopted to generate illumination weights to guide adaptive fusion of the inter- and intra-modality features.

Moreover, Zhang et al. \cite{Zhang25} devised an aligned region CNN (AR-CNN) to address the position shift problem between different modalities for multimodal object detection, where a region feature alignment module was constructed to calculate the position shift between VIS and IR modalities and assist in aligning cross-modal region features. To elevate the accuracy of position offset prediction, the AR-CNN introduces a region of interest (RoI) jitter strategy to facilitate sufficient feature fusion via feature reweighting. Li et al. \cite{Li25} designed a new frequency-driven feature decomposition network (FD$^2$-Net) to exploit frequency information of complementary features from different modalities for VIS-IR object detection, in which the high-frequency unit and low-frequency unit were used to capture high and low frequency features from two modalities, respectively. To facilitate multimodal feature fusion, a multimodal reconstruction mechanism in the FD$^2$-Net was developed to reduce image information loss.

Despite these advancements, these CNN-based detection methods cannot model long-range dependencies and exploit global feature information, which may lead to insufficient and incomplete cross-modal feature fusion, thereby interfering with multispectral object detection performance.

\subsection{Vision Transformer for Multispectral Object Detection}
Due to the outstanding performance of the transformer \cite{Vaswani17} in the field of natural language processing, transformer-based architecture is subsequently applied to the computer vision tasks \cite{Dosovitskiy21, Liu21, LiuH22}. Fang et al. \cite{Fang21} first introduced the transformer architecture into the field of multispectral object detection and achieved excellent performance, where a cross-modality fusion transformer (CFT) was designed to learn long-range relations and fuse global cross-modal feature information. Later, Xie et al. \cite{Xie23} introduced a feature interaction and self-attention fusion network for multispectral object detection, which utilized the feature interaction module to capture the correlations of different modalities by cross-modal information interaction, and applied a self-attention feature fusion module to model long-range dependencies to mine complementary features. You et al. \cite{You23} presented a multi-scale aggregation network (MSANet) for multispectral object detection, in which a multi-scale aggregation transformer was devised from the standard transformer \cite{Vaswani17} to interact with intermodality feature information, and a cross-modal merging fusion mechanism was developed to aggregate complementary information from VIS and IR modalities. Zhang et al. \cite{ZhangL24} devised a differential feature awareness network (DFANet) within antagonistic learning for VIS-IR object detection, in which the attention-based differential feature fusion module is adopted to integrate the differential multimodal features captured by the proposed antagonistic feature extraction with divergence module. Zeng et al. \cite{Zeng24} proposed a novel framework, namely MMI-Det, for VIS-IR object detection, where a contour enhancement module was constructed to enhance the object contours in VIS images, and a fusion focus module was used for sufficiently combining the object detail information from VIS and IR modalities. Dong et al. \cite{Dong25} developed a multimodal object detection approach, called SeaDate, which designed a dual attention feature fusion module consisting of a spatial multi-head attention and a channel group attention to capture spatial and channel feature information and effectively merge global and local features.

The above multispectral detection methods utilize the self-attention mechanism in the transformer to naturally extract complementary features between modalities, which can only capture limited complementary information. To this end, some multispectral detection approaches based on the cross attention mechanism have been proposed to better exploit cross-modal complementary features. Lee et al. \cite{Lee24} presented a cross-guided attention module to fuse inter-modal features from a two-stream backbone network by applying a parallel transformer. Shen et al. \cite{Shen24} developed a new feature fusion framework, namely ICAFusion, for multispectral object detection, where dual cross attention transformer was introduced to exchange global cross-modal feature information and extract complementary features between VIS and IR modalities. Yang et al. \cite{Yang25} devised a novel multidimensional fusion network (MMFN) to capture multi-modal features from channel, local, and global levels and fully integrate these features for multispectral object detection. Hu et al. \cite{HuH25} proposed an edge-guided illumination-aware interactive learning-based detector (EI$^2$Det) for VIS-IR object detection, which applies an edge-guided fusion module and an illumination-aware weighting module to improve detection performance by the obtained edge information and fuse cross-modal features by predicting the illumination intensity to weight multi-modal features, respectively.

However, these multispectral detection methods use all tokens to model long-range dependencies and capture complementary features between modalities, which introduces irrelevant feature information to calculate the similarity scores across modalities, resulting in redundant information generation and decreased detection performance.

\subsection{Sparse Mechanism}
Sparse mechanisms are widely used in computer vision fields to reduce computational cost or improve model performance. Chen et al. \cite{ChenL23} designed an efficient high-resolution vision transformer (SparseViT) for multiple computer vision tasks. Instead of applying structural sparsity to the attention mechanism, the SparseViT accelerates computation by leveraging activation sparsity. It scores the activation saliency of each attention window in the window-based visual transformer, prioritizing important windows and pruning unimportant ones. Zhu et al. \cite{Zhu21} proposed a deformable DETR for end-to-end object detection, where deformable attention does not perform global attention computation on all pixels, but learns a small set of key sampling points on the feature map for each query token and implements attention aggregation only at these locations to accelerate inference. Rao et al. \cite{Rao21} proposed a dynamic token sparsification framework (DynamicViT) that progressively prunes redundant tokens based on the input, in which a lightweight prediction module is devised to compute the importance score of each token and dynamically prunes tokens progressively across multiple layers. These pruned tokens no longer participate in attention computation, thereby reducing computational cost.

The above sparse studies aim to reduce computational complexity by collecting key points or pruned tokens for attention computation. Our sparse cross-attention mechanism uses a top-k strategy to explicitly suppress low weights between modalities and reduce interfering information, aiming to change the attention weights to improve detection performance.

\section{METHODOLOGY}\label{Sec_Methodology}
\subsection{The Proposed Framework}
\begin{figure*}[htbp]
	\begin{center}
		\includegraphics[width=0.90\textwidth]{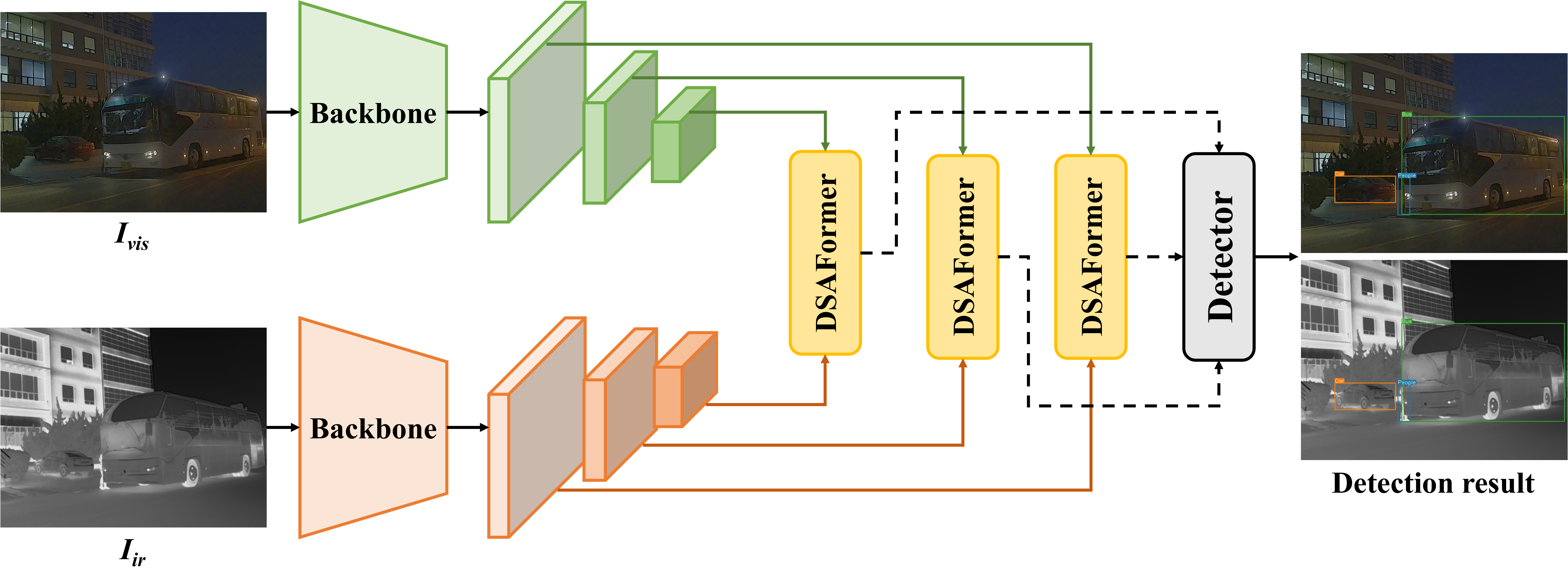}
		\caption{The proposed multispectral object detection framework, where the DSAFormer denotes our designed Dual Sparse Aggregation Transformer.}
        \label{fig1:Framework}
	\end{center}
\end{figure*}

Our proposed multispectral object detection framework is displayed in Fig. \ref{fig1:Framework}, which consists of a dual backbone network, three DSAFormers, and a detector, where a two-stream backbone network is used to extract multi-scale multimodal features from VIS and IR images, and the DSAFormer (Dual Sparse Aggregation Transformer) is proposed to sufficiently fuse multi-scale cross-modal features, and then the fused features are fed into the detector to generate the final detection results. The fusion and detection pipeline is formulated as follows:
\begin{align}
        P_{cls}, P_{box} = Det(DSAFormer(B(I_{vis}, I_{ir}))),
\end{align}
where $I_{vis}$ and $I_{ir}$ stand for visible light (VIS) and infrared (IR) images, and $B$ is the dual-branch backbone network. $P_{cls}$ and $P_{box}$ represent the classification categories and bounding boxes of the objects in VIS and IR images predicted by the detector $Det$.

\subsection{Dual Sparse Aggregation Transformer}
\begin{figure}[htbp]
	\begin{center}
		\includegraphics[width=0.45\textwidth]{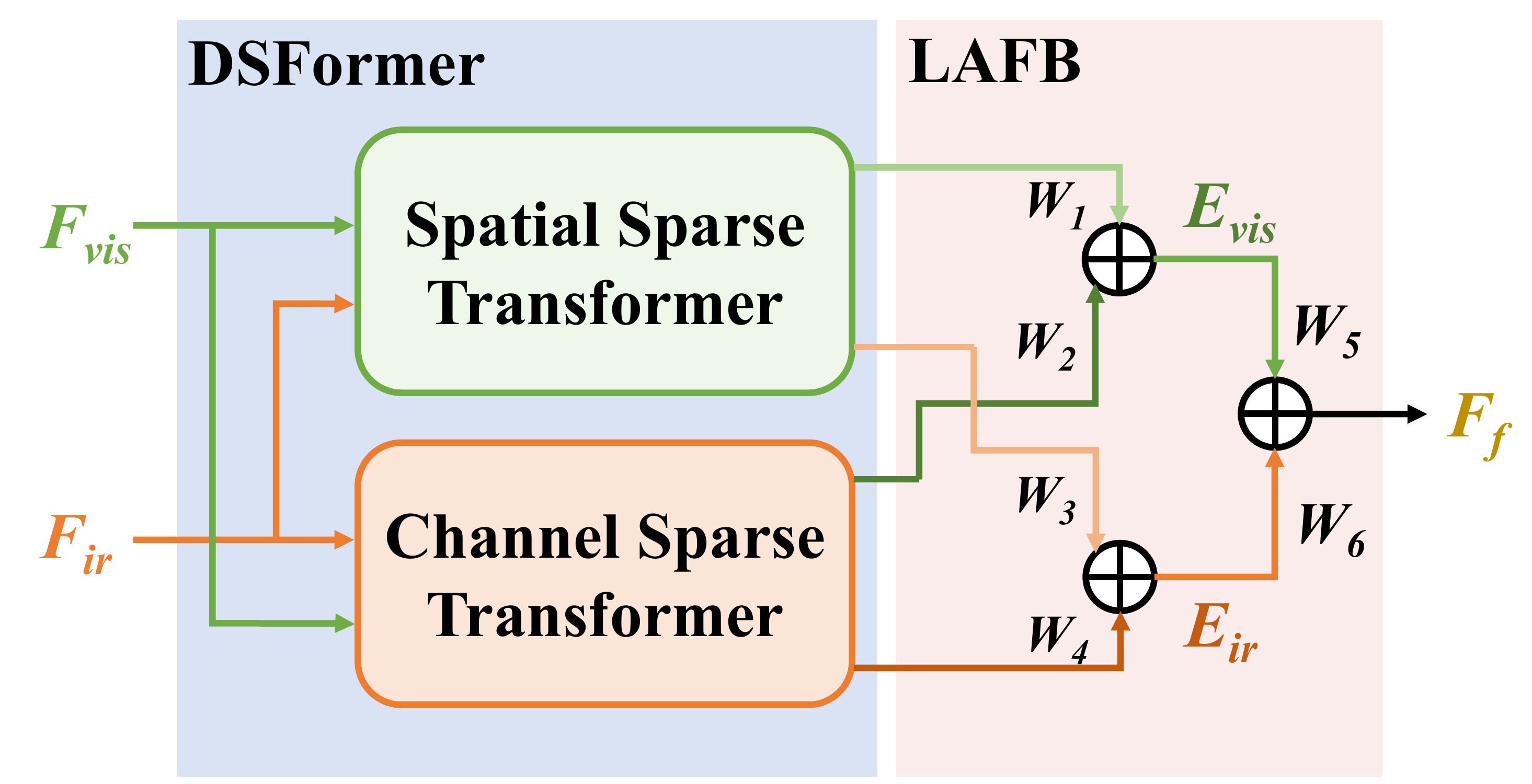}
		\caption{The architecture of the proposed Dual Sparse Aggregation Transformer (DSAFormer). The DSAFormer consists of a Dual Sparse Transformer (DSFormer) and a Learnable Addition Fusion Block (LAFB).}
        \label{fig2:DSAFormer}
	\end{center}
\end{figure}

To integrate spatial and channel information between VIS and IR modalities, the existing SeaDate \cite{Dong25} designs a dual-attention transformer to realize information interaction at the spatial and channel levels across modalities, but overlooks the redundant information generated by dense cross-modal exchanges. In order to achieve effective utilization of spatial and channel information between modalities and remove redundant information, a novel Dual Sparse Aggregation Transformer (DSAFormer) is proposed, as shown in Fig. \ref{fig2:DSAFormer}, which is applied to aggregate multi-scale spatial-channel features across modalities, where a Dual Sparse Transformer (DSFormer) is utilized to refine and enhance cross-modal complementary features, and its final outputs are combined by a Learnable Addition Fusion Block (LAFB) to produce appropriate fusion result by feature reweighting, whose procedure can be expressed as follows:
\begin{align}
    \begin{aligned}
        F_f = LAFB(DSFormer(F_{vis}, F_{ir}))
    \end{aligned}
\end{align}
where $F_{vis}$ and $F_{ir}$ stand for multimodal features extracted by the dual backbone network. $F_f$ is the final fusion feature, which is sent to the detector to obtain the detection results. The designed Dual Sparse Transformer (DSFormer) consists of a Spatial Sparse Transformer (SSFormer) and a Channel Sparse Transformer (CSFormer). The SSFormer and CSFormer are applied to exploit effective multimodal complementary information in the spatial and channel levels, respectively, where the mixed top-k selection mechanism is used to filter out low similarity scores between tokens from VIS and IR modalities. The LAFB contains multiple learnable feature weights, which are employed to merge two VIS and two IR features generated by the SSFormer and CSFormer separately, and are used to fuse the merged VIS and IR results by a learnable weight addition operation. The above process can be represented as follows:
\begin{align}
    \begin{aligned}
        E_{vis}^s, E_{ir}^s &= SSFormer(F_{vis}, F_{ir}), \\
        E_{vis}^c, E_{ir}^c &= CSFormer(F_{vis}, F_{ir}), \\
        E_{vis} &= W_{1}*E_{vis}^s + W_{2}*E_{vis}^c, \\
        E_{ir} &= W_{3}*E_{ir}^s + W_{4}*E_{ir}^c, \\
        F_f &= W_{5}*E_{vis} + W_{6}*E_{ir},
    \end{aligned}
\end{align}
where $E_{vis}^s$ and $E_{ir}^s$ are complementary features from the SSFormer with spatial information exchange, and $E_{vis}^c$ and $E_{ir}^c$ are complementary features from the CSFormer with channel feature interaction. These complementary features are enhanced and fused by the learnable weight addition operation `$\oplus$', where $W_{n}$ ($n$ $\in$ $\{1, 2, 3, 4, 5, 6\}$) is learnable weighting coefficient, which is initially set to 1.0.

\subsection{Spatial Sparse Transformer}\label{Sec_SSFormer}
\begin{figure*}[htbp]
	\begin{center}
		\includegraphics[width=0.90\textwidth]{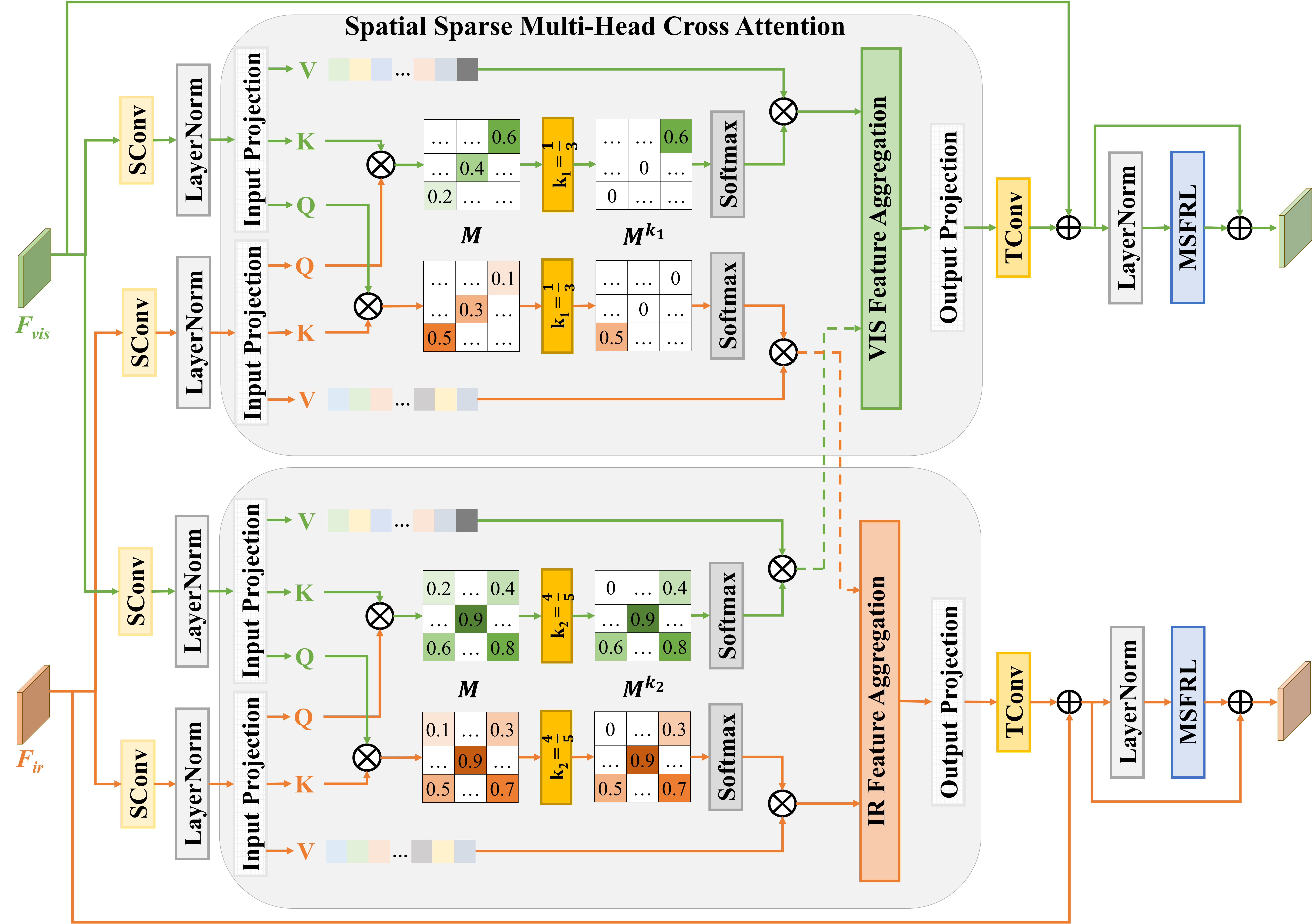}
		\caption{The structure of the proposed Spatial Sparse Transformer (SSFormer).}
        \label{fig3:SSFormer}
	\end{center}
\end{figure*}

Previous multimodal detection approaches \cite{Fang21}, \cite{You23}, \cite{Zeng24}, \cite{Yang25}, \cite{Zhu23} utilize spatial dense transformers (SDFormer) to exploit complementary features from VIS-IR modalities, where the attention maps are obtained by performing the matrix multiplication operation between all tokens from different modalities, which introduces irrelevant tokens into the similarity calculation, leading to noisy information interference.

To address this issue, we designed a Spatial Sparse Transformer (SSFormer) to mine effective spatial complementary features from VIS-IR modalities, which is shown in Fig. \ref{fig3:SSFormer}. The SSFormer mainly consists of four 2$\times$2 strided convolutions (SConv), two 2$\times$2 transposed convolutions (TConv), six Layer Normalization operations (LayerNorm), four residual connections, two Spatial Sparse Multi-Head Cross Attention (SSMHCA), and two Multi-Scale Feature Refinement Layers (MSFRL), whose overall process can be expressed as follows:
\begin{align}
    \begin{aligned}
        \bar{V}_{vis}^s, \bar{V}_{ir}^s &= SSMHCA(LN(SConv(F_{vis}, F_{ir}))), \\
        \bar{E}_{vis}^s &= TConv(\bar{V}_{vis}^s) + F_{vis}, \\
        \bar{E}_{ir}^s &= TConv(\bar{V}_{ir}^s) + F_{ir}, \\
        E_{vis}^s &= MSFRL(LN(\bar{E}_{vis}^s)) + \bar{E}_{vis}^s, \\
        E_{ir}^s &= MSFRL(LN(\bar{E}_{ir}^s)) + \bar{E}_{ir}^s,
    \end{aligned}
\end{align}
where LN denotes the Layer Normalization. $E_{vis}^s$ and $E_{ir}^s$ are the enhanced complementary features in the spatial level. To reduce the computational complexity of the SSMHCA, we apply the SConv to downsample the feature size to half of its original size and normalize the multimodal features by the LayerNorm to stabilize the model training process. Next, we utilize the SSMHCA to capture useful feature information by choosing the tokens with the top-k highest relevance between modalities, where the normalized multimodal features are projected to generate query ($Q_{vis}$, $Q_{ir}$), key ($K_{vis}$, $K_{ir}$), and value ($V_{vis}$, $V_{ir}$), whose sizes are all $d \times H \times W$. Then, the VIS attention matrix $M_{vis}^s$ $\in$ $R^{HW \times HW}$ is generated by the scaled dot-product operation between transposed $Q_{ir}$ and $K_{vis}$. Similarly, the IR attention matrix $M_{ir}^s$ $\in$ $R^{HW \times HW}$ is also obtained by the scaled dot-product operation between transposed $Q_{vis}$ and $K_{ir}$. These processes can be defined as follows:
\begin{align}
    \begin{aligned}
        M^s_{vis} &= \frac{Q_{ir}^{T}K_{vis}}{\lambda}, \\
        M^s_{ir} &= \frac{Q_{vis}^{T}K_{ir}}{\lambda},
    \end{aligned}
\end{align}
where $\lambda$ stands for the learnable scaling factor. Furthermore, we apply a mixed top-k selection strategy to filter out the low weights in $M_{vis}^s$ and $M_{ir}^s$, which enables a flexible setting process from dense attention maps to sparse attention maps. The sparse process requires a binary mask matrix $bmm^k$, which can be written as follows:
\begin{equation}\label{bmm}
bmm^k =
\begin{cases}
1, & p^{k} \in topk(M), \\
0, & otherwise.
\end{cases}
\end{equation}
Where $bmm^k$ is a binary mask matrix consisting of 0 and 1. $M$ represents the dense attention map, which can be either $M_{vis}^s$ or $M_{ir}^s$. $p^{k}$ denotes the position coordinates of the top-k highest weights in the attention map. Specifically, the $k$ value in the SSMHCA of Fig. \ref{fig3:SSFormer} is changing. For instance, when $k_{2} = \frac{4}{5}$, the best 80\% of the weights are reserved, and the remaining 20\% of the elements are set to 0. Thus, the sparse attention matrices are generated by the softmax function, which can be expressed as follows:
\begin{align}\label{sam}
    \begin{aligned}
        M^k_{vis} &= Softmax(bmm^k_{vis} \odot M^s_{vis}), \\
        M^k_{ir} &= Softmax(bmm^k_{ir} \odot M^s_{ir}),
    \end{aligned}
\end{align}
where $\odot$ stands for the element-wise product. $M^k_{vis}$ and $M^k_{ir}$ are the sparse attention matrixes with a sparse rate of $k$ in the SSMHCA. Subsequently, the matrix multiplication operation between the sparse attention maps ($M^k_{vis}$, $M^k_{ir}$) and the corresponding transposed values ($V_{vis}^T$, $V_{ir}^T$) is conducted to obtain multiple VIS and IR feature sparse representations, which can be formulated as follows:
\begin{align}\label{vsp}
    \begin{aligned}
        V^k_{vis} = M^k_{vis} \otimes V_{vis}^T, \\
        V^k_{ir} = M^k_{ir} \otimes V_{ir}^T,
    \end{aligned}
\end{align}
where $\otimes$ is the matrix multiplication operation. Since we use the mixed sparse rate $k_j$ ($k_j$ $\in$ [$\frac{1}{3}$, $\frac{4}{5}$], $j$ $\in$ $\{1, 2\}$) to retain different high matching query-key from VIS-IR modalities, the generated sparse representations $V_{vis}^{k_j}$ and $V_{ir}^{k_j}$ need to be aggregated for each of the 8 heads, as shown in Fig. \ref{fig3:SSFormer}, which can be denoted as follows:
\begin{align}\label{asp}
    \begin{aligned}
        \bar{V}_{vis}^i = \sum_{j=1}^{S} V_{vis}^{k_j} / S, \\
        \bar{V}_{ir}^i = \sum_{j=1}^{S} V_{ir}^{k_j} / S,
    \end{aligned}
\end{align}
where S is equal to 2 indicating two different values of $k$. $\bar{V}_{vis}^i$ and $\bar{V}_{ir}^i$ represent the VIS and IR average outputs of the $i$-th head in the SSMHCA, respectively, and $i$ ranges from 1 to 8 in increments of 1. As the multi-head scheme is adopted to enhance multimodal feature representation, we need to concatenate all the outputs of each multi-head sparse cross attention and obtain the final result by the output projection operation, which can be expressed as follows:
\begin{align}\label{pro}
    \begin{aligned}
        \bar{V}_{vis} = Projection(Concat(\bar{V}_{vis}^i)), \\
        \bar{V}_{ir} = Projection(Concat(\bar{V}_{ir}^i)),
    \end{aligned}
\end{align}
where $\bar{V}_{vis}$ and $\bar{V}_{ir}$ are the output of the Spatial Sparse Multi-Head Cross Attention (SSMHCA). In summary, the algorithm process of our proposed SSMHCA is shown in the Algorithm. \ref{alg:SSMHCA}. It's worth noting that the masking operations in this algorithm are utilized to generate the binary mask matrixes, as shown in Equation (6). Specifically, we employ the top-k algorithm to obtain the position information of the top k highest weights. Based on these positions, we set the top k highest weights to 1, while assigning 0 to the remaining lower weights. Thus, the binary mask matrixes composed of 0 and 1 are generated.

\begin{algorithm}[htbp]
\small
\caption{Spatial Sparse Multi-Head Cross Attention}
\setstretch{1.1}
\label{alg:SSMHCA}
\begin{algorithmic}[1]
\REQUIRE ~~\\
VIS feature $F_{vis}$ $\in$ $\mathbb{R}^{C \times H \times W}$, IR feature $F_{ir}$ $\in$ $\mathbb{R}^{C \times H \times W}$. \\
\ENSURE ~~\\
$f_{vis}^i$ $\in$ $\mathbb{R}^{d \times H \times W}$ and $f_{ir}^i$ $\in$ $\mathbb{R}^{d \times H \times W}$ are the VIS and IR features in $i$-th head, respectively, the total number of heads is equal to 8, $\lambda$ is the learnable temperature factor, $d$ = $\frac{C}{8}$, $k_j$ = [$\frac{1}{3}$, $\frac{4}{5}$], $M_{vis}^{k_j}$ and $M_{ir}^{k_j}$ are the zero matrix.
\vspace{2mm}
\FOR{each $f_{vis}^i$ and $f_{ir}^i$}
\STATE $Q_{vis}$, $K_{vis}$, $V_{vis}$ $\leftarrow{}$ Projection($f_{vis}^i$); \hfill // $d \times H \times W$ \\
\STATE $Q_{ir}$, $K_{ir}$, $V_{ir}$ $\leftarrow{}$ Projection($f_{ir}^i$); \hfill // $d \times H \times W$ \\
\STATE $Q_{vis}$, $K_{vis}$, $V_{vis}$ $\leftarrow{}$ Reshape($Q_{vis}$, $K_{vis}$, $V_{vis}$); \hfill // $d \times HW$ \\
\STATE $Q_{ir}$, $K_{ir}$, $V_{ir}$ $\leftarrow{}$ Reshape($Q_{ir}$, $K_{ir}$, $V_{ir}$); \hfill // $d \times HW$ \\
\STATE $M^s_{vis}$ $\leftarrow{}$ $\frac{Q_{ir}^T K_{vis}}{\lambda}$; \hfill // $HW \times HW$ \\
\STATE $M^s_{ir}$ $\leftarrow{}$ $\frac{Q_{vis}^T K_{ir}}{\lambda}$; \hfill // $HW \times HW$ \\
\FOR{each $k_j$}
\STATE $p_{vis}^{k_j}$ $\leftarrow{}$ topk($M^s_{vis}$); \\
\STATE $p_{ir}^{k_j}$ $\leftarrow{}$ topk($M^s_{ir}$); \\
\STATE $bmm_{vis}^{k_j}$ $\leftarrow{}$ mask($M_{vis}^{k_j}$, $p_{vis}^{k_j}$, 1); \\
\STATE $bmm_{ir}^{k_j}$ $\leftarrow{}$ mask($M_{ir}^{k_j}$, $p_{ir}^{k_j}$, 1); \\
\STATE $M_{vis}^{k_j}$ $\leftarrow{}$ Softmax($bmm_{vis}^{k_j}$ $\odot$ $M_{vis}$); \\
\STATE $M_{ir}^{k_j}$ $\leftarrow{}$ Softmax($bmm_{ir}^{k_j}$ $\odot$ $M_{ir}$); \\
\STATE $\bar{V}_{vis}^{k_j}$ $\leftarrow{}$ $M_{vis}^{k_j}$ $\otimes$ $V_{vis}^T$; \hfill // $HW \times d$ \\
\STATE $\bar{V}_{ir}^{k_j}$ $\leftarrow{}$ $M_{ir}^{k_j}$ $\otimes$ $V_{ir}^T$; \hfill // $HW \times d$ \\
\ENDFOR
\STATE $\bar{V}_{vis}^{i}$ $\leftarrow{}$ $\sum_{j=1}^{S}$ $\bar{V}_{vis}^{k_j}$ / $S$; \hfill // $HW \times d$ \\
\STATE $\bar{V}_{ir}^{i}$ $\leftarrow{}$ $\sum_{j=1}^{S}$ $\bar{V}_{ir}^{k_j}$ / $S$; \hfill // $HW \times d$ \\
\STATE $\bar{V}_{vis}^{i}$ $\leftarrow{}$ Reshape($\bar{V}_{vis}^{i}$); \hfill // $d \times H \times W$ \\
\STATE $\bar{V}_{ir}^{i}$ $\leftarrow{}$ Reshape($\bar{V}_{ir}^{i}$); \hfill // $d \times H \times W$ \\
\ENDFOR
\STATE $\bar{V}_{vis}$ $\leftarrow{}$ Projection(Concat($\bar{V}_{vis}^{i}$)); \hfill // $C \times H \times W$ \\
\STATE $\bar{V}_{ir}$ $\leftarrow{}$ Projection(Concat($\bar{V}_{ir}^{i}$)); \hfill // $C \times H \times W$ \\
\RETURN $\bar{V}_{vis}$, $\bar{V}_{ir}$;
\end{algorithmic}
\end{algorithm}

Additionally, the TConv operation is employed to recover the feature shape to its initial size, and the residual connection is used to alleviate the information loss caused by the downsampling operation and avoid vanishing or exploding gradients. In order to accelerate model convergence and improve model generalization ability, a layer normalization operation is used to normalize the feature distribution of each layer. Also, we design a Multi-Scale Feature Refinement Layer (MSFRL) to further eliminate redundant information and exploit multi-scale features, which can be shown in Fig. \ref{fig4:MSFRL}, and the details of the MSFRL will be described in Section \ref{Sec_MSFRL}.

\subsection{Channel Sparse Transformer}
Compared to existing channel dense transformers (CDFormer) \cite{HuH25}, \cite{Lee24}, \cite{Dong25}, \cite{Shen24} to capture complementary features from VIS-IR modalities in the channel level without considering redundant information interaction, our developed Channel Sparse Transformer (CSFormer) mines high-quality complementary features in the channel level between multimodal relevant tokens, which can reduce noise information. The structure of the CSFormer is similar with the SSFormer in Fig. \ref{fig3:SSFormer}, which mainly contains four 2$\times$2 strided convolutions (SConv), two 2$\times$2 transposed convolutions (TConv), six Layer Normalization operations (LayerNorm), four residual connections, two Channel Sparse Multi-Head Cross Attention (CSMHCA), and two Multi-Scale Feature Refinement Layers (MSFRL). The process of the CSFormer can be denoted as follows:
\begin{align}
    \begin{aligned}
        \bar{V}_{vis}^c, \bar{V}_{ir}^c &= CSMHCA(LN(SConv(F_{vis}, F_{ir}))), \\
        \bar{E}_{vis}^c &= TConv(\bar{V}_{vis}^c) + F_{vis}, \\
        \bar{E}_{ir}^c &= TConv(\bar{V}_{ir}^c) + F_{ir}, \\
        E_{vis}^c &= MSFRL(LN(\bar{E}_{vis}^c)) + \bar{E}_{vis}^c, \\
        E_{ir}^c &= MSFRL(LN(\bar{E}_{ir}^c)) + \bar{E}_{ir}^c,
    \end{aligned}
\end{align}
where LN represents the Layer Normalization operation, and $E_{vis}^c$ and $E_{ir}^c$ are the enhanced complementary features across modalities. The difference between CSFormer and SSFormer as described in Section \ref{Sec_SSFormer} is only the multi-head cross attention. The CSFormer utilizes the Channel Sparse Multi-Head Cross Attention (CSMHCA) to extract latent corrections across modalities in the channel level. Similarly, the algorithm flow of the CSMHCA can refer to the Algorithm. \ref{alg:SSMHCA}. It should be noted that the size of the generated dense attention matrixes ($M_{vis}^c$, $M_{ir}^c$) and sparse attention matrixes ($M_{vis}^{k_i}$, $M_{ir}^{k_i}$) is $d \times d$. The dense attention matrices are obtained by the following:
\begin{align}
    \begin{aligned}
        M^c_{vis} &= \frac{Q_{ir}K_{vis}^{T}}{\lambda}, \\
        M^c_{ir} &= \frac{Q_{vis}K_{ir}^{T}}{\lambda},
    \end{aligned}
\end{align}
where $\lambda$ is the learnable scaling factor. $M^c_{vis}$ and $M^c_{ir}$ stand for VIS and IR dense attention maps in the CSMHCA, respectively. The calculation process of the sparse attention matrixes, VIS-IR feature aggregation, and output projection in the CSMHCA are the same as those of the SSMHCA, as seen in Equations (\ref{bmm}), (\ref{sam}), (\ref{vsp}), (\ref{asp}), (\ref{pro}). It should be noted that the mixed sparsity rate $k_j$ ($k_j$ $\in$ [$\frac{2}{3}$, $\frac{4}{5}$], $j$ $\in$ $\{1, 2\}$) are adopted to suppress the noise information in the CSMHCA. In addition, the Layer Normalization, the Multi-Scale Feature Refinement Layer, and the residual connection are also applied to strengthen feature representation.

\subsection{Multi-Scale Feature Refinement Layer}\label{Sec_MSFRL}
\begin{figure}[htbp]
	\begin{center}
		\includegraphics[width=0.35\textwidth]{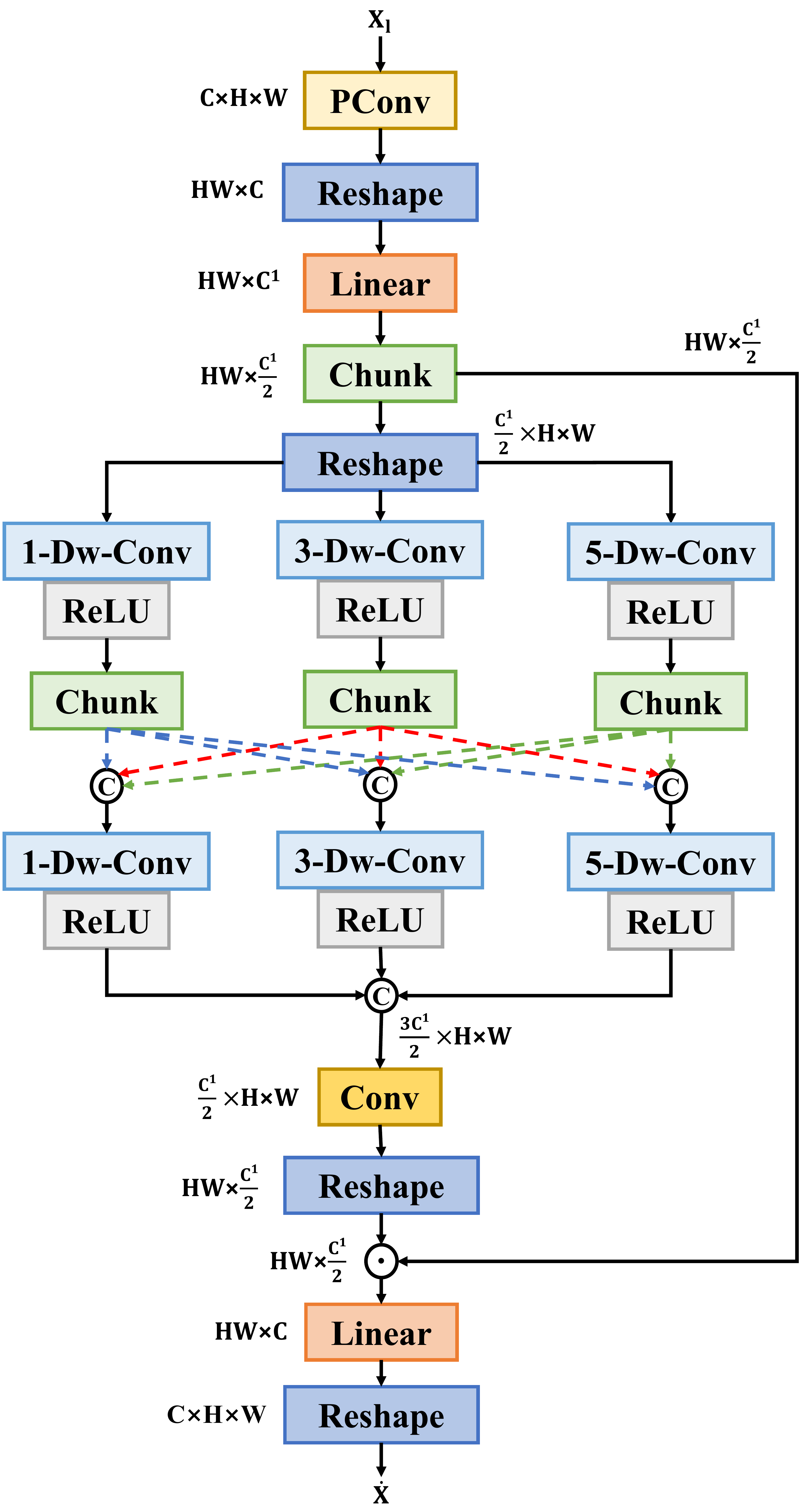}
		\caption{The structure of the proposed Multi-Scale Feature Refinement Layer (MSFRL).}
        \label{fig4:MSFRL}
	\end{center}
\end{figure}

The common FFN \cite{Vaswani17} utilizes two linear projections with a ReLU activation for feature propagation, but it does not exploit multi-scale features for feature representation, which limits its local feature enhancement. Furthermore, the SSMHCA and CSMHCA are used to remove noisy information by the mixed token selection scheme, but VIS and IR feature aggregation operations, as shown in Fig. \ref{fig3:SSFormer}, may introduce redundant information. To solve these drawbacks, we design a Multi-Scale Feature Refinement Layer (MSFRL) to capture multi-scale properties and eliminate irrelevant information, which can be displayed in Fig. \ref{fig4:MSFRL}. Specifically, a partial convolution (PConv)  \cite{Chen23} is applied to enhance the informative elements, and a reshape operation and a linear projection are used to adjust feature shape and expand feature dimension, which is formulated as follows:
\begin{align}
    \begin{aligned}
        X' = Linear(R(f_{pconv}^{3\times3}(X_l))), \\
    \end{aligned}
\end{align}
where $X_l$ is the normalized feature processed by a Layer Normalization operation, and $X'$ is the output of a linear layer. $f_{pconv}^{3\times3}$ and $R$ denote the $3 \times 3$ partial convolution and reshape operation, respectively. Next, a gate mechanism is adopted to decrease the redundant information, where a multi-scale representation is developed to explore multi-scale properties by using $1\times1$, $3\times3$, and $5\times5$ depth-wise convolutions (i-Dw-Conv, i $\in$ \{1, 3, 5\}) \cite{Howard17} with the ReLU activation function, which can be expressed as follows:
\begin{align}
    \begin{aligned}
        \hat{X}, \check{X} &= Chunk(X'), \\
        X_1 &= \sigma(f_{dwc}^{1\times1}(R(\hat{X}))),\\
        X_3 &= \sigma(f_{dwc}^{3\times3}(R(\hat{X}))),\\
        X_5 &= \sigma(f_{dwc}^{5\times5}(R(\hat{X}))),\\
    \end{aligned}
\end{align}
where $\hat{X}$ and $\check{X}$ are split features with the same size, which are generated by the chunk operation. $R$ and $\sigma$ denote the reshape operation and the ReLU activation function, respectively. $f_{dwc}^{1\times1}$, $f_{dwc}^{3\times3}$, and $f_{dwc}^{5\times5}$ stand for $1\times1$, $3\times3$, and $5\times5$ depth-wise convolution, respectively. $X_1$, $X_3$, and $X_5$ are the output features with multi-scale representation. To promote the information exchange, we chunk these features into three average patches in channel dimension, and apply the concatenation operation, depth-wise convolution with different kernel sizes, and ReLU function to merge these multi-scale features, which can be defined as follows:
\begin{align}
    \begin{aligned}
        \tilde{X}_1 &= \sigma(f_{dwc}^{1\times1}([X_1^{p1}, X_3^{p1}, X_5^{p1}])),\\
        \tilde{X}_3 &= \sigma(f_{dwc}^{3\times3}([X_1^{p2}, X_3^{p2}, X_5^{p2}])),\\
        \tilde{X}_5 &= \sigma(f_{dwc}^{5\times5}([X_1^{p3}, X_3^{p3}, X_5^{p3}])),\\
        \tilde{X} &= [\tilde{X}_1, \tilde{X}_3, \tilde{X}_5],\\
    \end{aligned}
\end{align}
where [,] is the concatenation operation, and $\tilde{X}$ is the output result of multi-scale representation. To restore the feature size before the multi-scale design, we adopt a $1 \times 1$ standard convolution (Conv) and a reshape operation to reduce the number of channels and adjust the feature shape. Moreover, an element-wise product is performed to decrease interference information. Finally, a linear layer and a reshape operation are employed to recover the initial feature dimension and shape. The above process can be denoted as follows:
\begin{align}
    \begin{aligned}
        \bar{X} &= R(f_{conv}^{1\times1}(\tilde{X})),\\
        \dot{X} &= R(Linear(\bar{X} \odot \check{X})),
    \end{aligned}
\end{align}
where $f_{conv}^{1\times1}$, $R$, and $\odot$ stand for the $1\times1$ standard convolution layer, the reshape operation, and the element-wise product, respectively. $\dot{X}$ is the enhanced output of the MSFRL.

\subsection{Loss Function}
The loss function of the designed multispectral object detection framework is mainly used to optimize model parameters to obtain more accurate object categories and locations, which can be summarized as the classification and regression tasks. The loss function consists of the classification loss ($\mathcal{L}_{cls}$), the confidence loss ($\mathcal{L}_{obj}$), and the regression loss ($\mathcal{L}_{box}$), which can be represented as:
\begin{equation}
    \mathcal{L} = \mathcal{L}_{cls} + \mathcal{L}_{obj} + \mathcal{L}_{box},
\end{equation}
where $\mathcal{L}_{cls}$ and $\mathcal{L}_{obj}$ apply the cross-entropy (CE) loss \cite{Boer05}, and $\mathcal{L}_{box}$ utilizes the generalized intersection over union (GIoU) loss \cite{Rezatofighi19}.

\section{EXPERIMENTS}\label{Sec_Experiments}
\subsection{Datasets}
To validate the effectiveness of our DSAFormer, we select the MFAD \cite{HuH25}, FLIR \cite{Zhang20}, M$^3$FD \cite{Liu22}, and LLVIP \cite{Jia21} datasets for experimental analysis. Due to the complementarity of VIS and IR modalities utilized by our method to enhance detection performance, these datasets are well aligned or slightly misaligned after image registration algorithms, and share a uniform label. Therefore, only a single multi-scale detection head is needed during model training and testing to detect the objects in both VIS and IR images. Moreover, these datasets have different resolutions and numbers of classes, fully verifying the robustness of our model.

\subsubsection{MFAD}
The MFAD dataset \cite{HuH25} is a well-aligned benchmark for multispectral object detection collected in scenes such as road, viaduct, tunnel, campus, and others with different illumination intensity levels, which consists of 12194 VIS-IR image pairs with a resolution of \text{$1280 \times 960$} pixels. 9751 and 2443 VIS-IR image pairs are employed to perform model training and assessment, individually. The dataset contains six categories, namely ``Car'', ``Bus'', ``Truck'', ``Pedestrian'', ``EbikeRider'', and ``Cyclist''.

\subsubsection{FLIR}
The FLIR dataset \cite{Zhang20} is a weakly misaligned benchmark, which is challenging for multispectral object detection. This dataset is composed of 5142 VIS-IR image pairs with a size of \text{$640 \times 520$} pixels, which are taken in daytime and nighttime traffic street scenes. 4129 and 1013 VIS-IR image pairs are applied for network training and testing, respectively. The dataset has three categories, namely ``person", ``car", and ``bicycle".

\subsubsection{M$^3$FD}
The M$^3$FD dataset \cite{Liu22} is a slightly misaligned multispectral object detection dataset filmed in complex traffic and occluded scenes during day and night, which contains 4200 VIS-IR image pairs with a size of \text{$1024 \times 768$} pixels. This dataset has not been publicly split into training and testing sets. Similar to \cite{Liang23}, 3360 and 840 VIS-IR image pairs are used for model training and evaluation, separately. The dataset consists of six categories, namely ``People", ``Car", ``Bus", ``Lamp", ``Motorcycle", and ``Truck".

\subsubsection{LLVIP}
The LLVIP dataset \cite{Jia21} is a large-scale well-aligned multispectral benchmark for pedestrian detection, which is captured in daytime and low-light traffic environments. This dataset is comprised of 15488 VIS-IR image pairs with a size of \text{$1280 \times 1024$} pixels. 12025 and 3463  VIS-IR image pairs are utilized for model training and testing, respectively. The dataset only includes one category, namely ``person".

\subsection{Experimental Settings}
Multi-head strategy is applied with 8 parallel heads in both SSMHCA and CSMHCA, and the selection rates in SSMHCA and CSMHCA are set to [$\frac{1}{3}$, $\frac{4}{5}$] and [$\frac{2}{3}$, $\frac{4}{5}$], respectively. The SGD optimizer is utilized to optimize the parameters of our DSAFormer with an initial learning rate $1.0 \times 10^{-2}$ and a momentum of 0.937, and the weight decay is set to $5.0 \times 10^{-4}$. Also, we adopt the warmup and cosine annealing strategy to control the learning rate decay. For all the used datasets, the input resolution of training images is adjusted to $640 \times 640$ pixels. In the test stage, the image size of all datasets is not modified. Data augmentation approaches such as random rotation and mosaic are used on these datasets. Following the state-of-the-art CFT \cite{Fang21}, YOLOFusion \cite{Fang22}, ICAFusion \cite{Shen24}, CrossFormer \cite{Lee24}, MMFN \cite{Yang25}, EI$^2$Det \cite{HuH25}, we also employ the CSPDarknet53 and YOLOv5 detection head to extract multi-scale multimodal features from VIS-IR images and predict the object categories and bounding boxes, respectively. The DSAFormer model is respectively trained in 150, 150, 200, and 100 epochs with a batch size of 4 for the MFAD, FLIR, M$^3$FD, and LLVIP datasets, ensuring that the DSAFormer model is adequately trained. We apply Python 3.9.0 and Pytorch 1.12.1 to train and test our proposed DSAFormer. All experiments are executed on the Ubuntu 20.04.1 system with a 16-core Intel(R) Xeon(R) Gold 6226R CPU @ 2.90 GHz and a single NVIDIA GeForce RTX 3090 GPU. The experimental setup remains unchanged unless otherwise specified.

\subsection{Evaluation Metric}
In this work, we choose the Average Precision (AP) of the common object detection metric as the evaluation criterion, which is widely used in
\cite{Fang21, ZhangY23, Cao23, Fang22, Xie23, Chen22, You23, Yang25, Lee24, Shen24, Dong25, Zeng24, HuH25}. We classify positive and negative samples based on the correctness of the object classification and the Intersection over Union (IoU) threshold. The Mean Average Precision (mAP) denotes the average result of the AP for all categories in each dataset. Specifically, the mAP50 value stands for the mean average precision at IoU=0.50, and the mAP50:95, abbreviated as mAP, represents the mean average precision at different IoU thresholds, whose range is from 0.50 to 0.95 with an increase value of 0.05. A higher mAP50 or mAP value indicates better performance of the detection model.

\subsection{Comparison with State-of-the-art Methods}
In this section, we conduct comparative experiments on detection performance with other state-of-the-art models on the MFAD \cite{HuH25}, FLIR \cite{Zhang20}, M$^3$FD \cite{Liu22}, and LLVIP \cite{Jia21} datasets, including quantitative and qualitative comparisons, which are shown in Tables \ref{Tab1_MFAD}-\ref{Tab4_LLVIP} and Figs. \ref{fig5:MFAD}, \ref{fig6:FLIR}, \ref{fig8:M3FD}.

\subsubsection{Experiment on the MFAD Dataset}

\begin{table}[htbp]
\centering
\caption{The quantitative comparison of our devised DSAFormer and other outstanding approaches on the MFAD dataset. The top values are emphasized in bold.}
\label{Tab1_MFAD}
\begin{tabular}{c|c|cc}
\hline
Methods & Modality & mAP50($\%$) & mAP($\%$) \\
\hline
YOLOv5 \cite{YOLOv5} & VIS & 74.9  & 49.1 \\

YOLOv5 \cite{YOLOv5} & IR & 70.0 & 42.8\\

YOLOv10 \cite{Wang24} & VIS & 71.1 & 48.9 \\

YOLOv10 \cite{Wang24} & IR & 65.7 & 41.8 \\
\hline
TarDAL \cite{Liu22} & VIS+IR & 69.8 & 43.9 \\

CFT \cite{Fang21} & VIS+IR & 77.8 & 52.5 \\

TINet \cite{ZhangY23} & VIS+IR & 69.1 & 43.6 \\

ICAFusion \cite{Shen24} & VIS+IR & 77.6 & 52.7 \\

MMI-Det \cite{Zeng24} & VIS+IR & 76.9 & 51.4 \\

Fusion-Mamba \cite{DongZ25} & VIS+IR & 76.5 & 51.6 \\

RSDet \cite{Zhao26} & VIS+IR & $\mathbf{79.3}$ & 53.2 \\

EI$^{2}$Det* \cite{HuH25} & VIS+IR & 76.3 & 50.8 \\

EI$^{2}$Det \cite{HuH25} & VIS+IR & 79.0 & 53.3 \\
\hline
DSAFormer(Ours) & VIS+IR & $\mathbf{79.3}$ & $\mathbf{54.3}$ \\
\hline
\end{tabular}
\end{table}

\begin{figure}[htbp]
	\begin{center}
		\includegraphics[width=0.45\textwidth]{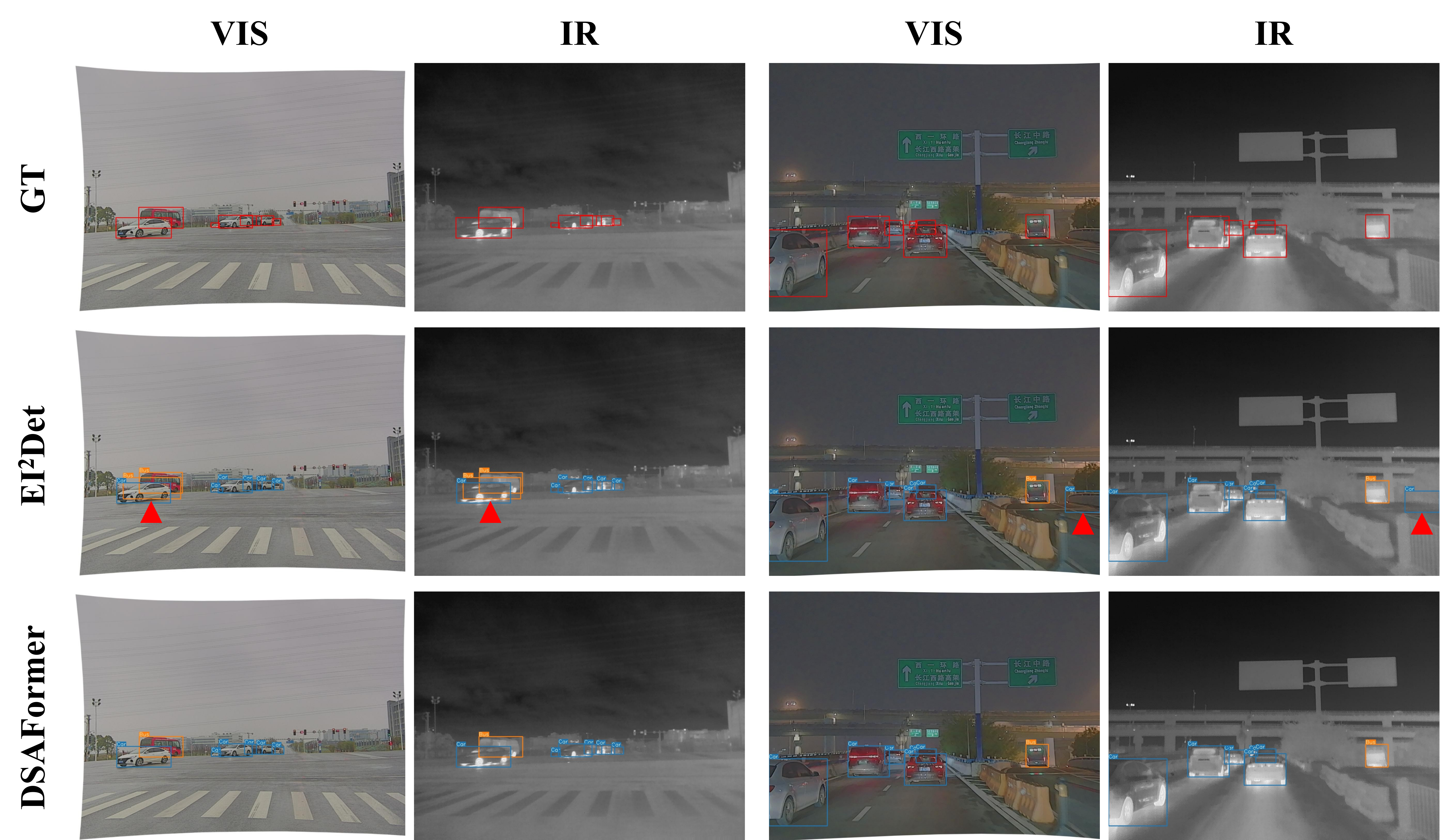}
		\caption{The detection samples of multispectral object detection methods on the MFAD dataset. The red triangles point to incorrect prediction results.}
        \label{fig5:MFAD}
	\end{center}
\end{figure}

In Table \ref{Tab1_MFAD}, we display the performance comparison of different detection approaches across VIS and IR modalities on the MFAD dataset. The top results are emphasized in bold. For monomodal detection methods YOLOv5 and YOLOv10, the mAP50 values are equal to 74.9\% and 71.1\% for VIS modality, and 70.0\% and 65.7\% for IR modality. The mAP values are 49.1\% and 48.9\% for VIS modality, and 42.8\% and 41.8\% for IR modality. It is evident that YOLOv5 performs better than YOLOv10 on the MFAD dataset, which suggests that more advanced general object detectors may not necessarily perform better in the field of multispectral object detection. Moreover, the detection results of these single modal detection methods are inferior to those of multimodal fusion detection methods such as CFT, ICAFusion, MMI-Det, EI$^2$Det, and our proposed DSAFormer, which proves that fusing multimodal complementary information can effectively improve detection performance. In the mAP metric, our DSAFormer outperforms the better monomodal detection methods YOLOv5 by 5.2\% and 11.5\% across VIS and IR modalities, respectively. Further, the DSAFormer also surpasses other state-of-the-art multispectral detection methods CFT, ICAFusion, MMI-Det, and EI$^2$Det by 1.8\%, 1.6\%, 2.9\%, and 1.0\%, respectively. In addition to category and bounding box labels, the MFAD dataset also takes the light intensity of VIS images as an extra label. The EI$^2$Det method uses this prior knowledge to improve multimodal detection performance, while our DSAFormer method does not adopt this prior and still achieves better detection performance than it, which fully demonstrates the effectiveness of our method. In addition, the EI$^2$Det without the illumination-aware weighting module, called EI$^2$Det*, which means that the light intensity of VIS images cannot be used, is employed for experimental comparison, as shown in Table \ref{Tab1_MFAD}.

In order to show our detection effect more intuitively, we visualize the detection results of two pairs of VIS-IR images from the MFAD dataset, as seen in Fig. \ref{fig5:MFAD}, where the ground-truth (GT) boxes and the prediction boxes generated by the EI$^2$Det and our DSAFormer are used to compare, and the red triangles mark detection errors. From Fig. \ref{fig5:MFAD}, we can see that the EI$^2$Det generates error detection boxes in the two pairs of VIS-IR images. The EI$^2$Det predicts an extra ``Bus" category in the left VIS-IR image pair, and incorrectly generates the ``Car'' class in the right VIS-IR image pair. However, our DSAFormer produces accurate prediction boxes and class names.

\subsubsection{Experiment on the FLIR Dataset}

\begin{table}[htbp]
\centering
\caption{The mAP50 and mAP values of our proposed DSAFormer and other attractive models on the FLIR dataset. The top results are highlighted in bold.}
\label{Tab2_FLIR}
\begin{tabular}{cccc}
\hline
Methods & Modality & mAP50($\%$) & mAP($\%$)\\
\hline
Faster R-CNN \cite{Ren17} &  VIS & 65.0 & 30.2\\

Faster R-CNN \cite{Ren17}  & IR & 73.4  & 37.9\\

YOLOv5 \cite{YOLOv5} & VIS & 67.8 & 31.2\\

YOLOv5 \cite{YOLOv5} & IR & 74.4 & 38.0\\
\hline
GAFF \cite{Zhang21} & VIS+IR & 72.9 & 37.5\\

CFT \cite{Fang21} & VIS+IR & 78.3 & 40.2\\

CSAA \cite{Cao23}  & VIS+IR  & 79.2 & 41.3 \\

YOLOFusion \cite{Fang22} & VIS+IR & 76.6 & 39.8\\

YOLO-MS \cite{Xie23} & VIS+IR & 75.3 & 38.3\\

ProbEn \cite{Chen22} & VIS+IR & 75.5 & 37.9\\

MSANet \cite{You23} & VIS+IR & 76.2 & 39.0\\

MMFN \cite{Yang25} & VIS+IR & 80.8 & 41.7\\

CrossFormer \cite{Lee24} & VIS+IR  & 79.3 & $\mathbf{42.1}$ \\

ICAFusion \cite{Shen24} & VIS+IR & 79.2 & 41.4\\

SeaDate \cite{Dong25} & VIS+IR & 80.3 & 41.3\\

MMI-Det \cite{Zeng24} & VIS+IR & 79.8 & 40.5\\

FD$^2$-Net \cite{Li25} & VIS+IR & 82.9 & -\\

DPAL \cite{Liu25} & VIS+IR & 76.0 & -\\

DWSF-Net \cite{Fan26} & VIS+IR & 82.3  & 41.8\\

EI$^2$Det \cite{HuH25} & VIS+IR & 80.2  & -\\
\hline
DSAFormer(Ours) & VIS+IR & $\mathbf{83.5}$  & 42.0 \\
\hline
\end{tabular}
\end{table}

\begin{figure*}[htbp]
	\begin{center}
		\includegraphics[width=0.90\textwidth]{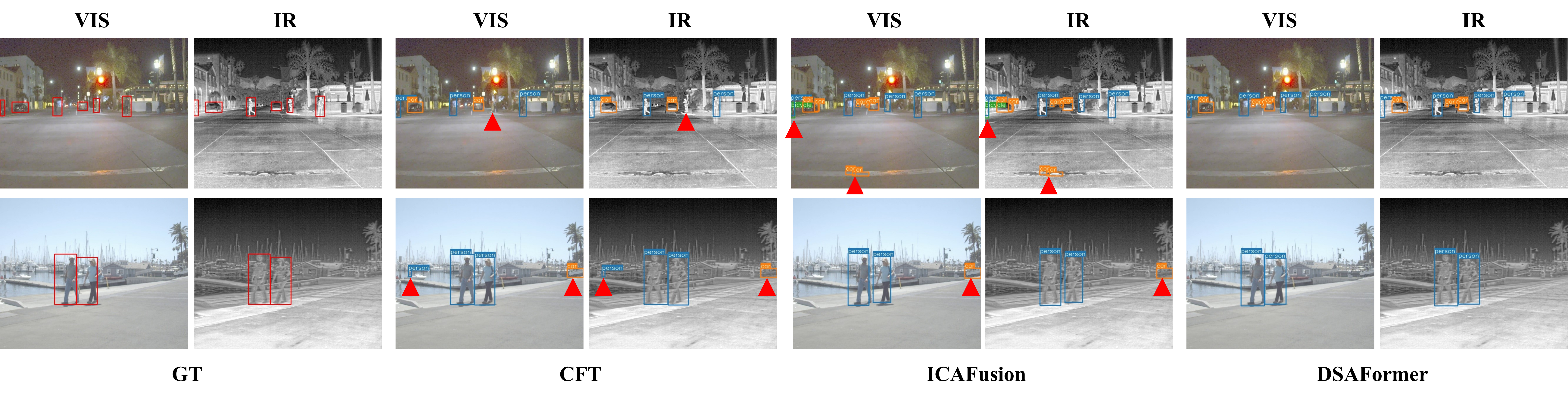}
		\caption{The qualitative comparison of multispectral object detection methods on the FLIR dataset. The red triangles mark missing or error prediction boxes.}
        \label{fig6:FLIR}
	\end{center}
\end{figure*}

In Table \ref{Tab2_FLIR}, we provide a quantitative comparison between our designed DSAFormer and other advanced detection methods on the FLIR dataset. The top results are highlighted in bold. For monomodal detection approaches Faster R-CNN and YOLOv5, the mAP50 values are 65.0\% and 67.8\% for VIS modality, and 73.4\% and 74.4\% for IR modality, respectively. These values are significantly lower than the 83.5\% mAP50 of our DSAFormer. Additionally, compared with thirteen multispectral detection methods such as CFT, MMFN, CrossFormer, ICAFusion, SeaDate, MMI-Det, EI$^2$Det, and so on, our DSAFormer obtains excellent detection performance in the mAP50 and mAP values. It should be noted that the CrossFormer is slightly ahead of our DSAFormer in terms of mAP metric, but its model complexity is higher. Subsequently, we conduct a visualization experiment on two pairs of VIS-IR images from the FLIR dataset, which can be shown in Fig. \ref{fig6:FLIR}. The first row of the image pair is taken from nighttime traffic scenes, where the CFT miss the boxes of ``person'' category, and the ICAFusion generate incorrect prediction boxes. The image pair in the second row is in a daytime scene, where the CFT and ICAFusion produce error prediction boxes. For example, they predict the redundant ``car'' category. In summary, our DSAFormer can accurately predict the object category and locate the object area.

\begin{figure}[htbp]
	\begin{center}
		\includegraphics[width=0.45\textwidth]{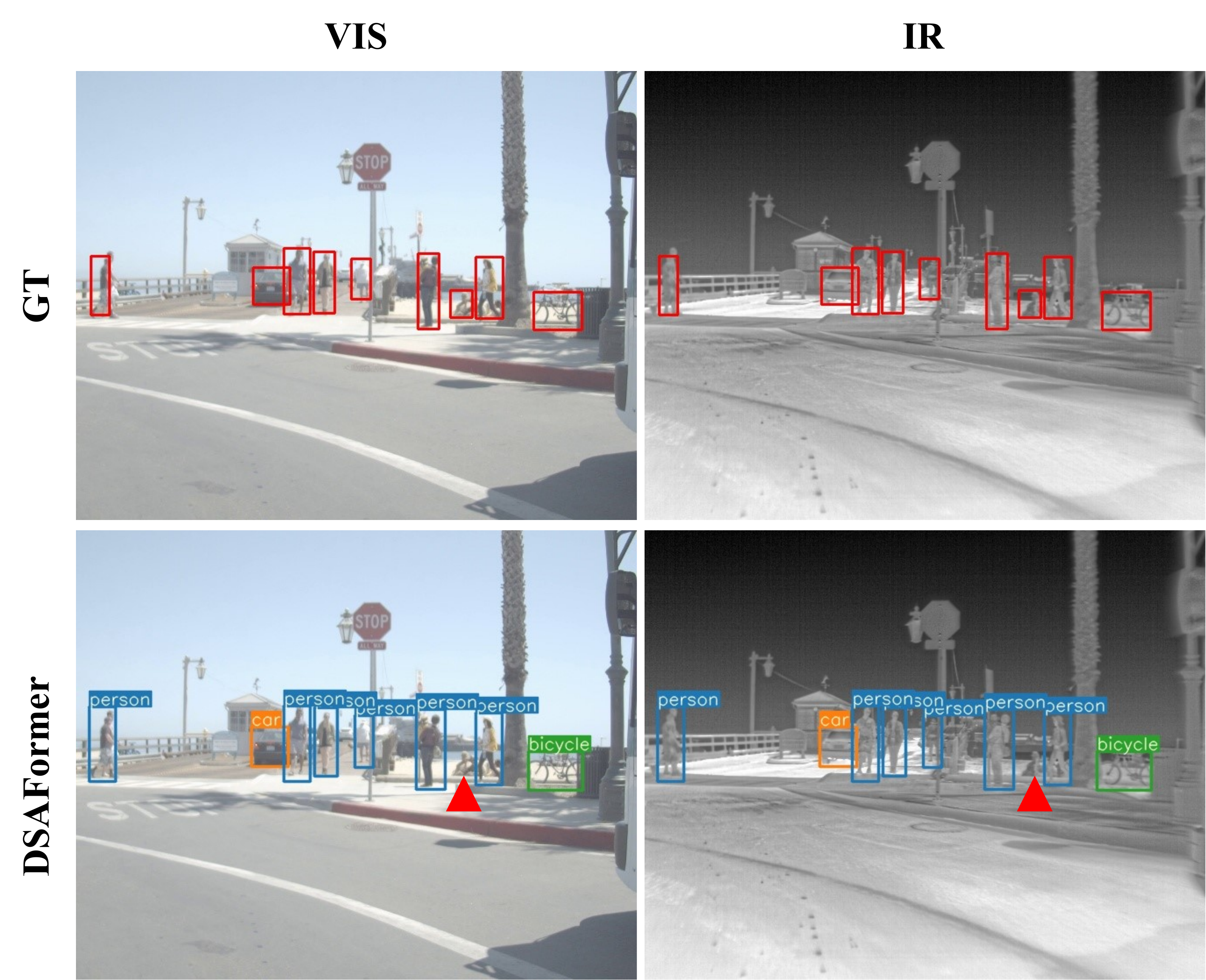}
		\caption{The failed detection case of our DSAFormer on a pair of VIS and IR images from the FLIR dataset. The red triangles mark missing prediction boxes.}
        \label{fig7:Failure}
	\end{center}
\end{figure}

In Fig. \ref{fig7:Failure}, our DSAFormer model fails to detect the ``sitting person''. Specifically, the prior appearance of the person category in the FLIR dataset is highly concentrated in the standing posture. During the learning process, the DSAFormer model uses upright contours and complete limb proportions as strong discriminative features. When the object is seated, its scale, aspect ratio, and key structures change significantly, causing a deviation from the learned category distribution and resulting in missed detections. Furthermore, a seated person tends to blend with the background (chair, ground, etc.) in thermal radiation, leading to incomplete human contours in the IR modality. The significant differences between VIS and IR in this posture further exacerbate the difficulty of aligning cross-modal features. Ultimately, under the dual influence of pose diversity modeling and insufficient cross-modal robustness, our DSAFormer struggles to correctly generalize the ``sitting person'' to the ``person'' category.

\subsubsection{Experiment on the M$^3$FD Dataset}

\begin{table}[htbp]
\centering
\caption{The performance comparison of our designed DSAFormer and other excellent methods on the M$^3$FD dataset. The best values are underlined in bold.}
\label{Tab3_M3FD}
\begin{tabular}{cccc}
\hline
Methods & Modality & mAP50($\%$) & mAP($\%$)\\
\hline
YOLOv5 \cite{YOLOv5} & VIS & 73.8 & 42.9\\

YOLOv5 \cite{YOLOv5} & IR & 68.6 & 40.4\\

YOLOv8 \cite{YOLOv8} & VIS & 80.9 & 52.5\\

YOLOv8 \cite{YOLOv8} & IR & 79.5 & 53.1\\
\hline
TarDAL \cite{Liu22} & VIS+IR & 80.5 & 54.1\\

DIDFuse \cite{Zhao20} & VIS+IR & 78.9 & 52.6\\

SDNet \cite{ZhangM21} & VIS+IR & 79.0 & 52.9\\

CFT \cite{Fang21} & VIS+IR & 88.2 & 59.0 \\

DeFusion \cite{SunC22} & VIS+IR & 80.8 & 53.8\\

SuperFusion \cite{TangD22} & VIS+IR & 83.5 & 56.0\\

RFNet \cite{Xu22} & VIS+IR & 79.4 & 53.2\\

CDDFuse \cite{ZhaoB23} & VIS+IR & 81.1 & 54.3\\

IGNet \cite{Li23} & VIS+IR & 81.5 & 54.5\\

ICAFusion \cite{Shen24} & VIS+IR & 87.1 & 57.0 \\

MS2Fusion \cite{Shen26} & VIS+IR & 89.4 & 59.7 \\

Fusion-Mamba \cite{DongZ25} & VIS+IR & 85.0 & 57.5 \\
\hline
DSAFormer(Ours) & VIS+IR & $\mathbf{89.7}$  & $\mathbf{60.4}$ \\
\hline
\end{tabular}
\end{table}

\begin{figure*}[htbp]
	\begin{center}
		\includegraphics[width=0.90\textwidth]{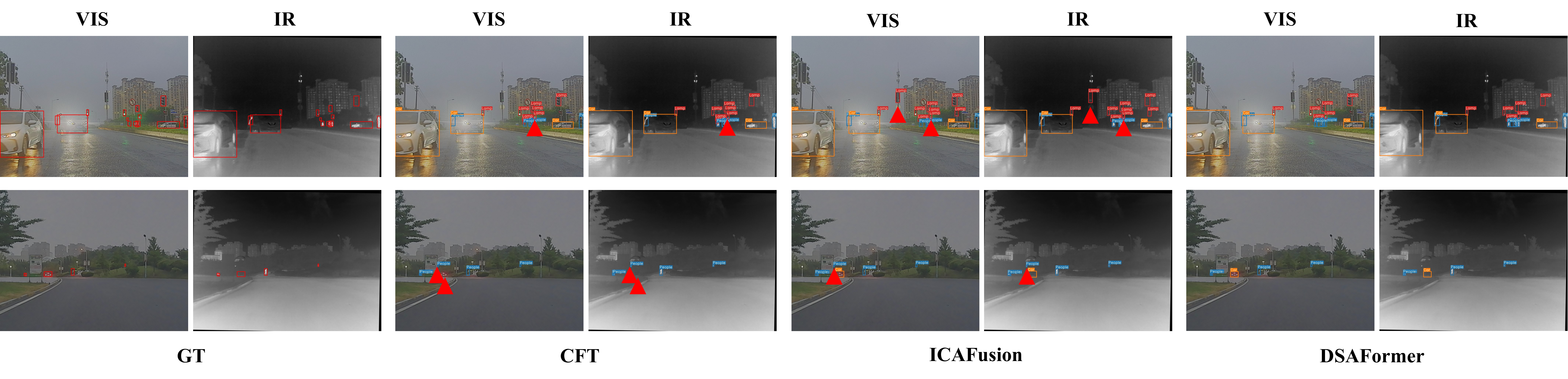}
		\caption{The visual comparison of multispectral object detection models on the M$^3$FD dataset. The red triangles represent missing or erroneous prediction boxes.}
        \label{fig8:M3FD}
	\end{center}
\end{figure*}

In Table \ref{Tab3_M3FD}, we list a numerical comparison between our developed DSAFormer and other detection models on the M$^3$FD dataset. The best mAP50 and mAP values are underlined in bold. For single spectrum detection methods YOLOv5 and YOLOv8, the mAP values are 42.9\% and 52.5\% for VIS modality, and 40.4\% and 53.1\% for IR modality, respectively. These results are inferior to the 60.4\% mAP obtained by our DSAFormer. Also, the DSAFormer outperforms multispectral object detection approaches such as CFT, ICAFusion, and so on. In conclusion, our method achieves excellent detection performance on the M$^3$FD dataset compared with other advanced detection models.

In addition, we conduct a visual comparison experiment on two pairs of VIS-IR images from the M$^3$FD dataset, which is displayed in Fig. \ref{fig8:M3FD}. We can see that the CFT and ICAFusion generate incorrect bounding boxes for the first row of VIS-IR images, which are marked by the red triangles. For the second row of VIS-IR images, the CFT produces a redundant prediction box with the ``People'' category and misses a bounding box of the ``Car'' class, and the ICAFusion predicts an incorrect bounding box with the ``People'' category, where the red triangles are used to point to missing and incorrect bounding boxes. However, our DSAFormer can predict the correct bounding boxes and object categories for these two pairs of VIS-IR images.

\subsubsection{Experiment on the LLVIP Dataset}
\begin{table}[htbp]
\centering
\caption{The detection results of our DSAFormer and other competitive methods on the LLVIP dataset. The best results are highlighted in bold.}
\label{Tab4_LLVIP}
\begin{tabular}{cccc}
\hline
Methods & Modality & mAP50($\%$) & mAP($\%$)\\
\hline
Faster R-CNN \cite{Ren17} & VIS & 88.8 & 47.5 \\

Faster R-CNN \cite{Ren17} & IR & 92.6 & 50.7\\

YOLOv5 \cite{YOLOv5} & VIS & 90.8 & 50.0\\

YOLOv5 \cite{YOLOv5} & IR & 94.6 & 61.9\\
\hline
GAFF \cite{Zhang21} & VIS+IR & 94.0 & 55.8\\

CFT \cite{Fang21} & VIS+IR & 97.5 & 63.6\\

CSAA \cite{Cao23} & VIS+IR & 94.3 & 59.2\\

YOLOFusion \cite{Fang22} & VIS+IR & 95.4 & 62.8\\

TINet \cite{ZhangY23} & VIS+IR & 94.3 & 60.6\\

ProbEn \cite{Chen22} & VIS+IR & 93.4 & 51.5\\

CrossFormer \cite{Lee24} & VIS+IR  & 97.5 & 65.1\\

TarDAL \cite{Liu22} & VIS+IR & 94.9 & 61.2\\

MMFN \cite{Yang25} & VIS+IR & 97.2 & -\\

ICAFusion \cite{Shen24} & VIS+IR & 96.3 & 62.3\\

CCLDet \cite{Shang25} & VIS+IR & 96.5 & -\\

RSDet \cite{Zhao26} & VIS+IR & 95.8 & 61.3\\

UniRGB-IR \cite{Yaun25} & VIS+IR & 96.1 & 63.2\\

Fusion-Mamba \cite{DongZ25} & VIS+IR & 96.8 & 62.8\\

FD$^2$-Net \cite{Li25} & VIS+IR & 96.2 & -\\

EI$^2$Det \cite{HuH25} & VIS+IR & $\mathbf{98.0}$ & 63.9\\
\hline
DSAFormer(Ours) & VIS+IR & 97.2 & $\mathbf{65.3}$ \\
\hline
\end{tabular}
\end{table}

In Table \ref{Tab4_LLVIP}, we report a comparison of the mAP50 and mAP values for different detection methods on the LLVIP dataset. Since this dataset is well-aligned and has only one ``person'' category, the detection results of different models are relatively great, especially for the mAP50 metric. The best results are highlighted in bold. Specifically, in terms of comprehensive metric mAP, monomodal object detection models Faster R-CNN and YOLOv5 obtain values of 47.5\% and 50.0\% for VIS modality, and 50.7\% and 61.9\% for IR modality, respectively. These values are lower than 65.3\% of the proposed DSAFormer. Besides, our method performs better on the mAP metric than other multispectral detection models. Although our DSAFormer is lower than the CFT, CrossFormer, and EI$^2$Det in the mAP50 value, it outperforms these methods in the comprehensive mAP metric, which still reflects the advantages of our method.

\subsection{Ablation Study}

\begin{table}[htbp]
\centering
\caption{The detection performance and testing time of different $k$ values in the Spatial Transformer (SFormer) on the M$^3$FD dataset. The most excellent results are marked in bold.}
\label{Tab5_ablation}
\begin{tabular}{c|c|ccc}
\hline
\multicolumn{2}{c|}{k value} & mAP50 & mAP & Time\\
\hline
\multirow{7}*{Fixed k} & 1 & 88.7$\%$ & 59.8$\%$ & 15.2ms\\
\cline{2-5}
  & $\frac{1}{6}$ & 88.6$\%$ & 59.8$\%$ & 15.5ms\\
\cline{2-5}
  & $\frac{1}{3}$ & 89.1$\%$ & 60.0$\%$ & 16.3ms\\
\cline{2-5}
  & $\frac{1}{2}$ & 89.1$\%$ & 59.8$\%$ & 16.2ms\\
\cline{2-5}
  & $\frac{2}{3}$ & 88.7$\%$ & 60.0$\%$ & 15.9ms\\
\cline{2-5}
  & $\frac{3}{4}$ & 88.9$\%$ & 60.1$\%$ & 16.2ms\\
\cline{2-5}
  & $\frac{4}{5}$ & 89.0$\%$ & 60.0$\%$ & 16.8ms\\
\cline{1-5}
\multirow{6}*{Mixed k} & [$\frac{1}{3}$, $\frac{1}{2}$] & 88.9$\%$ & 59.9$\%$ & 18.6ms\\
\cline{2-5}
  & [$\frac{1}{3}$, $\frac{3}{4}$] & $\mathbf{89.3}\%$ & 59.9$\%$ & 18.9ms\\
\cline{2-5}
  & [$\frac{1}{3}$, $\frac{4}{5}$] & $\mathbf{89.3}\%$ & $\mathbf{60.4}\%$ & 19.2ms\\
\cline{2-5}
  & [$\frac{3}{4}$, $\frac{4}{5}$] & 88.8$\%$ & 60.0$\%$ & 19.6ms\\
\cline{2-5}
  & [$\frac{1}{2}$, $\frac{2}{3}$, $\frac{3}{4}$, $\frac{4}{5}$] & $\mathbf{89.3}\%$ & 60.1$\%$ & 25.5ms\\
\cline{2-5}
  & [$\frac{1}{3}$, $\frac{1}{2}$, $\frac{2}{3}$, $\frac{3}{4}$, $\frac{4}{5}$] & 89.1$\%$ & 60.0$\%$ & 28.1ms\\
\hline
\end{tabular}
\end{table}

\begin{table}[htbp]
\centering
\caption{The detection performance and inference time of different $k$ values in the Channel Sparse Transformer (CSFormer) on the M$^3$FD dataset. The largest values of each mAP are highlighted in bold.}
\label{Tab6_ablation}
\begin{tabular}{c|c|ccc}
\hline
\multicolumn{2}{c|}{k value} & mAP50 & mAP & Time \\
\hline
\multirow{7}*{Fixed k} & 1 & 88.5$\%$ & 59.8$\%$ & 15.0ms \\
\cline{2-5}
  & $\frac{1}{6}$ & 88.3$\%$ & 59.8$\%$ & 15.3ms \\
\cline{2-5}
  & $\frac{1}{3}$ & 88.8$\%$ & 60.0$\%$ & 15.0ms \\
\cline{2-5}
  & $\frac{1}{2}$ & 88.9$\%$ & 60.0$\%$ & 15.6ms \\
\cline{2-5}
  & $\frac{2}{3}$ & 88.9$\%$ & 60.1$\%$ & 14.8ms \\
\cline{2-5}
  & $\frac{3}{4}$ & 88.7$\%$ & 60.2$\%$ & 15.7ms \\
\cline{2-5}
  & $\frac{4}{5}$ & 88.8$\%$ & 60.1$\%$ & 15.1ms \\
\cline{1-5}
\multirow{6}*{Mixed k} & [$\frac{1}{2}$, $\frac{2}{3}$] & 89.0$\%$ & 60.0$\%$ & 15.7ms \\
\cline{2-5}
  & [$\frac{2}{3}$, $\frac{3}{4}$] & 89.1$\%$ & 60.1$\%$ & 15.9ms \\
\cline{2-5}
  & [$\frac{2}{3}$, $\frac{4}{5}$] & 89.1$\%$ & $\mathbf{60.3}\%$ & 15.3ms \\
\cline{2-5}
  & [$\frac{3}{4}$, $\frac{4}{5}$] & 89.1$\%$ & 60.2$\%$ & 15.6ms \\
\cline{2-5}
  & [$\frac{1}{2}$, $\frac{2}{3}$, $\frac{3}{4}$, $\frac{4}{5}$] & $\mathbf{89.2}\%$ & 60.2$\%$ & 15.7ms \\
\cline{2-5}
  & [$\frac{1}{3}$, $\frac{1}{2}$, $\frac{2}{3}$, $\frac{3}{4}$, $\frac{4}{5}$] & 89.0$\%$ & 60.2$\%$ & 16.6ms \\
\cline{2-5}
\hline
\end{tabular}
\end{table}

\begin{table}[htbp]
\centering
\caption{The detection performance comparison of the Dual Dense Transformer and our DSAFormer on the FLIR dataset. The top results are underlined in bold.}
\label{Tab7_ablation}
\begin{tabular}{ccc}
\hline
Metric & mAP50($\%$) & mAP($\%$) \\
\hline
Dual Dense Transformer & 81.9 & 41.2 \\

DSAFormer(Ours) & $\mathbf{83.5}$ & $\mathbf{42.0}$ \\
\hline
\end{tabular}
\end{table}

\begin{table}[htbp]
\centering
\caption{The detection performance comparison of the Spatial Dense Transformer, Channel Dense Transformer, Dual Dense Transformer, Spatial Sparse Transformer, Channel Sparse Transformer, and our DSAFormer on the M$^3$FD dataset. The top results are underlined in bold.}
\label{Tab8_ablation}
\begin{tabular}{ccc}
\hline
Metric & mAP50($\%$) & mAP($\%$) \\
\hline
Spatial Dense Transformer & 88.7 & 59.8 \\

Channel Dense Transformer & 88.5 & 59.8 \\

Dual Dense Transformer & 89.1 & 60.2 \\

Spatial Sparse Transformer & 89.3 & $\mathbf{60.4}$ \\

Channel Sparse Transformer & 89.1 & 60.3 \\

DSAFormer(Ours) & $\mathbf{89.7}$ & $\mathbf{60.4}$ \\
\hline
\end{tabular}
\end{table}

\begin{figure*}[htbp]
	\begin{center}
		\includegraphics[width=0.90\textwidth]{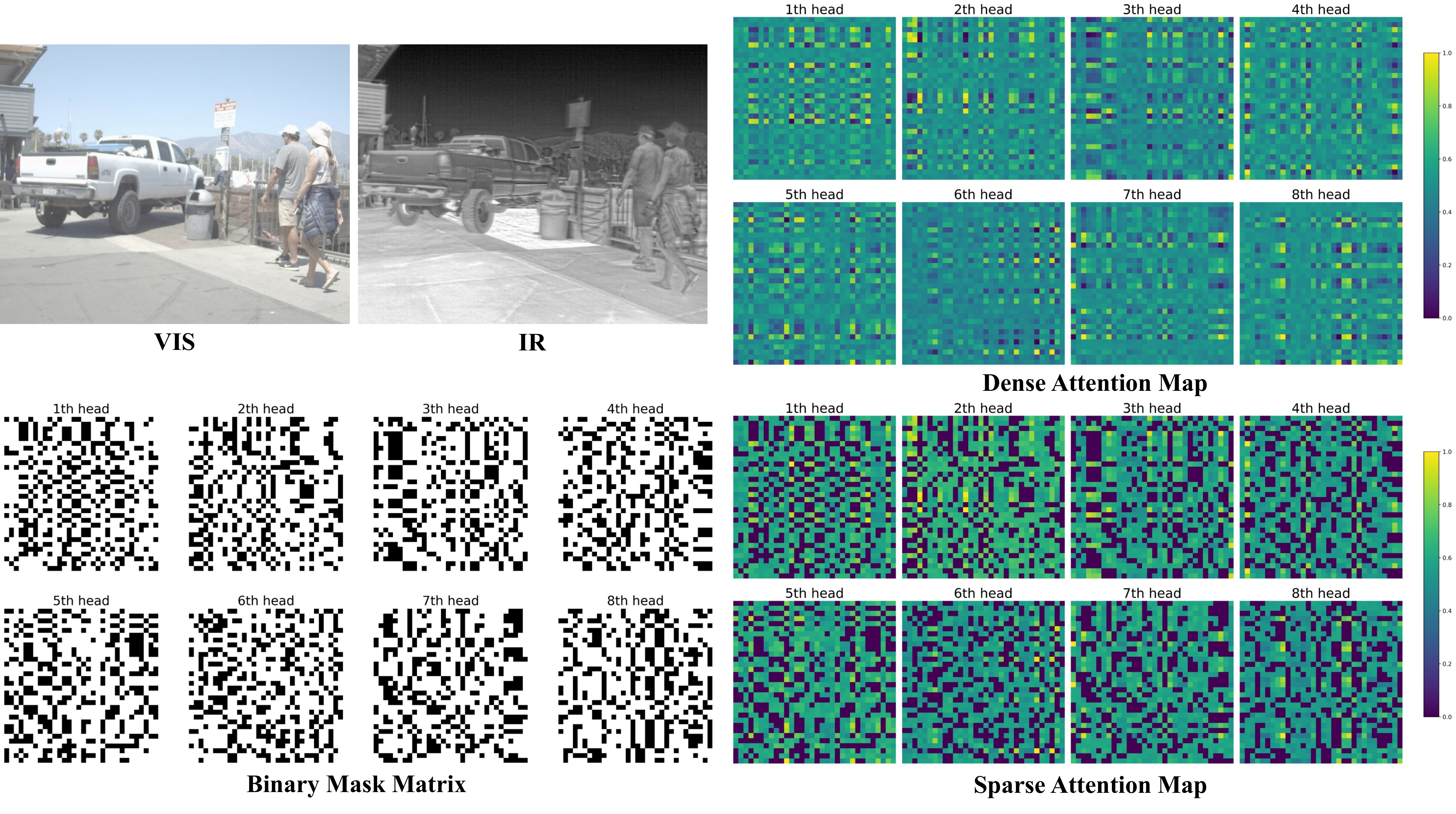}
		\caption{The visualization of multi-head dense and sparse attention maps and binary mask matrix for a pair of VIS-IR images from the FLIR dataset.}
        \label{fig9:Attention}
	\end{center}
\end{figure*}

\begin{table}[htbp]
\centering
\caption{The performance comparison of the MLP and MSFRL on the FLIR and M$^3$FD datasets, where the FFN, MLP, and MSFRL denote the Feed-Forward Network, Multilayer Perceptron, and Multi-Scale Feature Refinement Layer, respectively. The better results are marked in bold.}
\label{Tab9_MLP}
\begin{tabular}{c|c|cc}
\hline
Datasets & FFN & mAP50($\%$) & mAP($\%$)\\
\hline
\multirow{2}*{FLIR} & MLP & 81.3 & 41.2 \\

& MSFRL & $\mathbf{83.5}$ & $\mathbf{42.0}$ \\

\hline

\multirow{2}*{M$^3$FD} & MLP & 88.7 & 59.8\\

& MSFRL & $\mathbf{89.7}$ & $\mathbf{60.4}$ \\
\hline
\end{tabular}
\end{table}

\begin{table}[htbp]
\centering
\caption{The performance comparison of the DASFormer without($\times$) or with($\surd$) LAFB on the FLIR and M$^3$FD datasets, where the LAFB represents the Learnable Addition Fusion Block. The better values are highlighted in bold.}
\label{Tab10_LAFB}
\begin{tabular}{c|c|cc}
\hline
Datasets & LAFB & mAP50($\%$) & mAP($\%$)\\
\hline
\multirow{2}*{FLIR} & $\times$ & 83.1 & 41.3 \\

& $\surd$ & $\mathbf{83.5}$ & $\mathbf{42.0}$ \\
\hline
\multirow{2}*{M$^3$FD} & $\times$ & 89.4 & 60.3 \\

& $\surd$ & $\mathbf{89.7}$ & $\mathbf{60.4}$ \\

\hline
\end{tabular}
\end{table}

\begin{figure}[htbp]
	\begin{center}
		\includegraphics[width=0.45\textwidth]{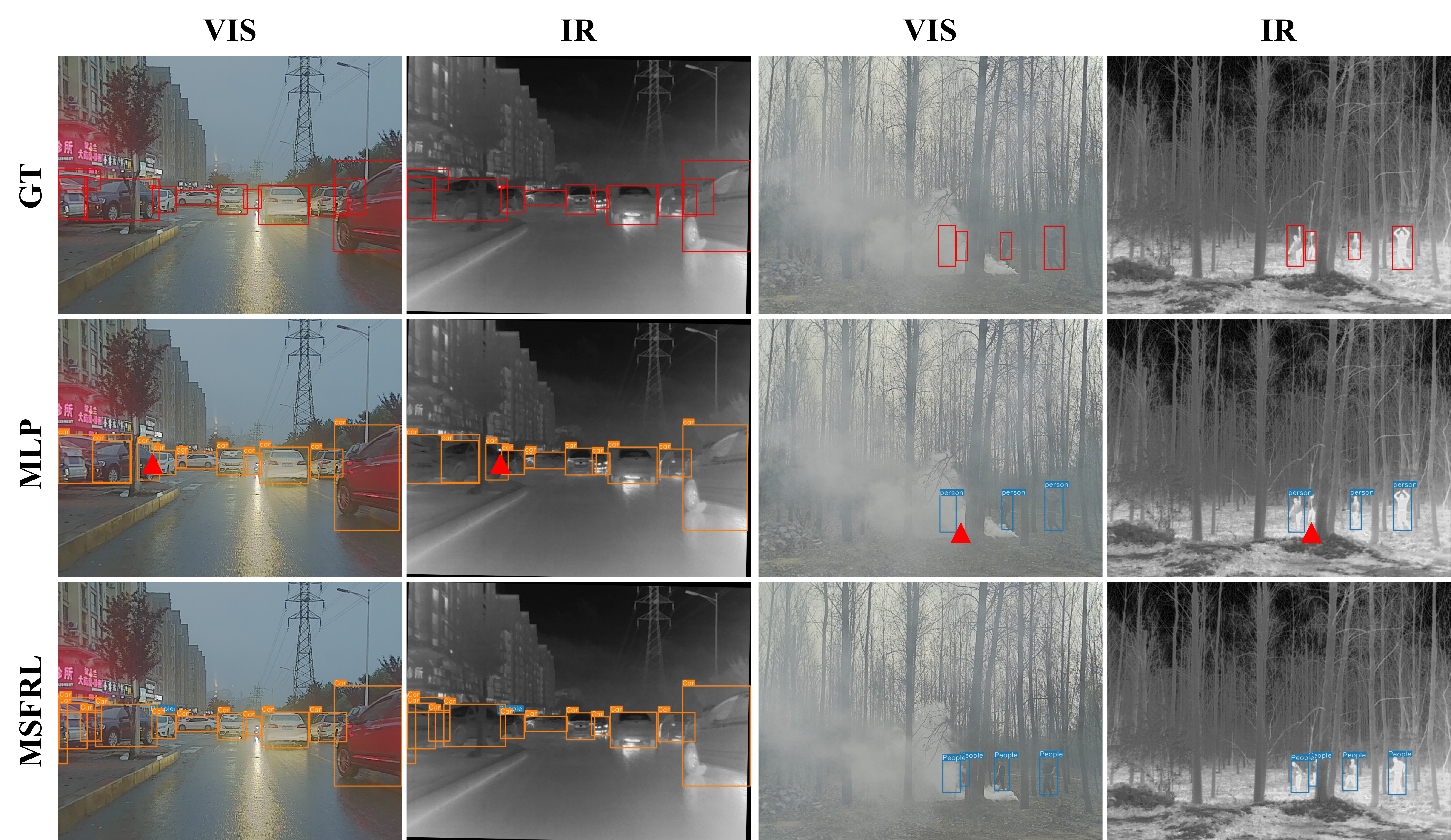}
		\caption{The visual comparison of MLP and MSFRL in different scenarios on the M$^3$FD dataset. The red triangles represent missing prediction boxes.}
        \label{fig10:FFN}
	\end{center}
\end{figure}

\begin{table}[htbp]
\centering
\caption{The performance comparison of different fusion strategies on the FLIR dataset. The better values are highlighted in bold.}
\label{Tab11_fusion}
\begin{tabular}{c|cc}
\hline
Fusion strategies & mAP50($\%$) & mAP($\%$)\\
\hline
Concatenation & 82.1 & 41.5 \\

Summation & 83.1 & 41.3\\

Gating  & 80.6 & 41.6\\

LAFB & $\mathbf{83.5}$ & $\mathbf{42.0}$\\
\hline
\end{tabular}
\end{table}

\begin{table}[htbp]
\centering
\caption{The ablation comparison of the SSFormer, CSFormer, MSFRL, and LAFB on the FLIR dataset. The top results are underlined in bold.}
\label{Tab12_ablation}
\begin{tabular}{ccccc}
\hline
SSFormer & CSFormer & MSFRL & LAFB & mAP50($\%$) \\
\hline
$\times$ & $\times$ & $\times$ & $\times$ & 79.8 \\

$\surd$ & $\times$ & $\times$ & $\times$ & 80.3 \\

$\times$ & $\surd$ & $\times$ & $\times$ & 80.7 \\

$\surd$ & $\surd$ & $\times$ & $\times$ & 81.2 \\

$\surd$ & $\times$ & $\surd$ & $\times$ & 81.6 \\

$\times$ & $\surd$ & $\surd$ & $\times$ & 81.3 \\

$\surd$ & $\surd$ & $\times$ & $\surd$ & 81.3 \\

$\surd$ & $\surd$ & $\surd$ & $\times$ & 83.1 \\

$\surd$ & $\surd$ & $\surd$ & $\surd$ & $\mathbf{83.5}$\\
\hline
\end{tabular}
\end{table}

\begin{table}[htbp]
\centering
\caption{The performance comparison of cross-dataset generalization testing. The better values are highlighted in bold.}
\label{Tab13_domain}
\begin{tabular}{c|ccc}
\hline
Dataset & \multicolumn{3}{c}{Target}\\
\hline
Source  & \multicolumn{3}{c}{LLVIP}\\
\hline
\multirow{4}*{LLVIP} & Methods & mAP50($\%$) & mAP($\%$)\\
\cline{2-4}
 & CFT \cite{Fang21} & $\mathbf{97.5}$ & 63.6 \\

 & ICAFusion \cite{Shen24} & 96.3 & 62.3 \\

 & DSAFormer & 97.2 & $\mathbf{65.3}$ \\
\hline
\multirow{3}*{FLIR}  & CFT \cite{Fang21} & 46.5 (-51.0) & 23.0 (-40.6) \\

 & ICAFusion \cite{Shen24} & 76.4 (-19.9) & 33.8 (-28.5) \\

 & DSAFormer & $\mathbf{77.6}$ (-$\mathbf{19.6}$) & $\mathbf{38.2}$ (-$\mathbf{27.1}$) \\
\hline
\end{tabular}
\end{table}

To verify the effectiveness of our DSAFormer, we conduct several ablation comparison experiments. In Tables \ref{Tab5_ablation} and \ref{Tab6_ablation}, we perform in-depth comparative experiments on the M$^3$FD dataset for different k values, where each model is trained for 200 epochs to ensure sufficient training and convergence. Furthermore, we control the random seed to avoid detection performance fluctuations due to initial randomness. Specifically, in Table \ref{Tab5_ablation}, we show the mAP50 and mAP values for different $k$ in the Spatial Transformer, where the M$^3$FD dataset containing multiple object categories and complex scenes is used for testing. When $k=1$, it means that the attention map is dense and the weights with low similarity are not removed. In addition, when $k$ is equal to $\frac{1}{6}$, $\frac{1}{3}$, $\frac{1}{2}$, $\frac{2}{3}$, $\frac{3}{4}$ or $\frac{4}{5}$, the attention matrix here is sparse with a fixed sparsity rate, and we eliminate low-weight elements of the matrix according to the value of $k$. Further, we mix different $k$ values to evaluate the performance of the model. From Table \ref{Tab5_ablation}, we can find that taking into account both mAP values and testing time, the mixed sparse strategy ($k$ $\in$ [$\frac{1}{3}$, $\frac{4}{5}$]) achieves the best mAP results, and the testing time also is competitive with other sparse schemes. Similarly, we display the mAP values and inference time for different $k$ in the Channel Transformer in Table \ref{Tab6_ablation}, and we select the unfixed sparse strategy ($k$ $\in$ [$\frac{2}{3}$, $\frac{4}{5}$]) as our baseline channel sparse transformer. Tables \ref{Tab5_ablation} and \ref{Tab6_ablation} can prove that our proposed sparse strategy using a mixed top-k selection mechanism is effective for multispectral object detection. From a practical application perspective, fixed and mixed sparsity strategies exhibit different trade-offs between performance improvement and computational complexity for our DSAFormer model. In offline multispectral object detection or scenarios with relatively abundant computing resources, the accuracy gains offered by mixed sparsity are reasonable and acceptable. However, in scenarios with strict real-time constraints, a fixed sparsity strategy remains more practical due to its simpler structure and stable inference. Therefore, the choice of sparsity strategy should be comprehensively evaluated based on the specific application scenario, computing budget, and real-time requirements.

Additionally, we demonstrate the effect of aggregating complementary information in the spatial and channel levels, where a quantitative comparison experiment is designed in Tables \ref{Tab7_ablation} and \ref{Tab8_ablation}. It can be seen from Table \ref{Tab8_ablation} that our developed DSAFormer produces the best mAP50 and mAP values compared with the Spatial Dense Transformer, Channel Dense Transformer, Dual Dense Transformer, Spatial Sparse Transformer, and Channel Sparse Transformer, which illustrates that combining the Spatial Sparse Transformer and Channel Sparse Transformer to integrate complementary spatial and channel information can improve the detection performance. The above findings are also supported by Table \ref{Tab7_ablation}. To better understand the sparse mechanism of our DSAFormer, we visualize the multi-head dense and sparse attention maps, as well as the binary mask matrix in the channel sparse multi-head cross attention, as illustrated in Fig. \ref{fig9:Attention}, where the sparsity rate $k$ is equal to $\frac{1}{3}$ and a pair of VIS-IR images are from the FLIR dataset. From Fig. \ref{fig9:Attention}, we can observe the process of filtering out low-weight values in the multi-head dense attention map to produce the multi-head sparse attention map. Moreover, to prove the effectiveness of our designed MSFRL, we perform a numerical comparison on the FLIR and M$^3$FD datasets, which is shown in Table \ref{Tab9_MLP}, where the Multilayer Perceptron (MLP) and our MSFRL are used as Feed-Forward Network (FFN) to compare. The better results are marked in bold. One can see that our MSFRL can achieve better detection performance. Also, we visualize the detection cases of MLP and MSFRL under different scenarios in Fig. \ref{fig10:FFN}. It can be seen that for the first image pair, the MLP fails to detect a small and occluded ``People'' object in the traffic scene, and generates a missed occluded ``People'' in the dense smoke environment in the second image pair. However, our MSFRL can achieve accurate localization in these complex scenarios, which demonstrates its advantages and effectiveness. Additionally, we explore the effectiveness of the proposed Learnable Addition Fusion Block (LAFB) on the FLIR and M$^3$FD datasets, in which an ablation comparison experiment is constructed in Table \ref{Tab10_LAFB}. The symbol `$\times$' from Table \ref{Tab10_LAFB} represents not using LAFB, so we remove six learnable weighting coefficients in the LAFB, including $W_1$, $W_2$, $W_3$, $W_4$, $W_5$, and use three simple summation operations to fuse multimodal features from the DSFormer. Further, the symbol `$\surd$' denotes the use of LAFB. It can be found that our designed LAFB can improve the detection performance. Furthermore, we compare the detection performance of the LAFB and other common fusion strategies on the FLIR dataset, as shown in Table \ref{Tab11_fusion}. It can be seen that our LAFB achieves better detection results. To verify the effectiveness of each proposed module, we conduct an ablation experiment on the FLIR dataset, as shown in Table \ref{Tab12_ablation}.  It can be seen that each module improves the detection performance, demonstrating their effectiveness and indispensability.

Subsequently, we conduct a cross-domain performance comparison experiment to evaluate the generalization ability of our DSAFormer method in comparison to other methods, as illustrated in Table \ref{Tab13_domain}. When both the source and target domains are the LLVIP dataset, our DSAFormer achieves competitive detection performance. Moreover, we train the DSAFormer model using the FLIR dataset as the source domain and then test its detection performance on the LLVIP target domain. Table \ref{Tab13_domain} reveals that our DSAFormer obtains better detection results. Notably, our DSAFormer exhibits the least performance degradation when the source and target domains are inconsistent, demonstrating its stronger generalization ability compared to other models.

\subsection{Complexity Comparison}

\begin{table}[htbp]
\centering
\caption{The complexity comparison of our DSAFormer and other state-of-the-art approaches, in which the resolution size of VIS-IR modalities is set to $640\times640$.}
\label{Tab14_complexity}
\begin{tabular}{cccc}
\hline
Methods & Modality & Param. & FLOPs \\
\hline
CFT \cite{Fang21} & VIS+IR & 206.0M & 224.6G \\

ProbEn \cite{Chen22} & VIS+IR & 945.9M & - \\

MSANet \cite{You23} & VIS+IR & 447.6M & 485.6G\\

SeaDate \cite{Dong25} & VIS+IR & 306.6M & 134.3G \\

MMFN \cite{Yang25} & VIS+IR & 176.4M & - \\

ICAFusion \cite{Shen24} & VIS+IR & 120.2M & 192.7G \\

EI$^2$Det \cite{HuH25} & VIS+IR & 127.7M & 220.9G \\

MMI-Det \cite{Zeng24} & VIS+IR & 207.6M & 229.2G \\

CrossFormer \cite{Lee24} & VIS+IR & 340.0M & 361.7G \\

MS2Fusion \cite{Shen26} & VIS+IR & 130.3M & 140.8G \\

Fusion-Mamba \cite{DongZ25} & VIS+IR & 223.7M & 466.1G \\

DSAFormer(Ours) & VIS+IR & 138.8M & 359.9G \\
\hline
\end{tabular}
\end{table}

In Table \ref{Tab14_complexity}, we conduct a complexity comparison experiment for our devised DSAFormer and other state-of-the-art detection methods, where the CFT, ProbEn, MSANet, SeaDate, MMFN, ICAFusion, EI$^2$Det, MMI-Det, CrossFormer, MS2Fusion, and Fusion-Mamba are used for comparison. From Table \ref{Tab14_complexity}, we can conclude that the number of parameters (Param.) of the DSAFormer is higher than that of the ICAFusion, EI$^2$Det, and MS2Fusion, but less than that of the eight remaining methods. Moreover, in terms of the floating point operations per second (FLOPs), the DSAFormer is higher than the CFT, SeaDate, ICAFusion, EI$^2$Det, MMI-Det, and MS2Fusion, but it is lower than the MSANet, CrossFormer, and Fusion-Mamba. However, our DSAFormer still has a relatively large number of FLOPs. The attention maps generated in the sparse attention mechanisms remove low weights using a mixed top-k selection strategy to eliminate redundant information from irrelevant token interactions between modalities. During this process, the attention weights are sorted, resulting in additional computational cost. In summary, compared to these excellent multispectral object detection methods, the computational complexity of our DSAFormer is competitive.

\section{CONCLUSION}\label{Sec_Conclusion}
In this work, we design a Dual Sparse Aggregation Transformer (DSAFormer) to achieve comprehensive multimodal feature integration for multispectral object detection. The core component of our framework, the Dual Sparse Transformer (DSFormer), is developed to capture and strengthen intermodal complementary features. Specifically, the DSFormer consists of a Spatial Sparse Transformer (SSFormer) and a Channel Sparse Transformer (CSFormer), which uses a Spatial Sparse Multi-Head Cross Attention (SSMHCA) and a Channel Sparse Multi-Head Cross Attention (CSMHCA) to exploit effective multimodal correlations in spatial and channel levels by utilizing the mixed top-k token selection strategy to choose high query-key matching scores from VIS-IR modalities, respectively. Moreover, the Multi-Scale Feature Refinement Layer (MSFRL) is applied in the SSFormer and CSFormer to extract multi-scale relations and further decrease redundant feature information. To sufficiently fuse multimodal features, a Learnable Addition Fusion Block (LAFB) is adopted to enhance and merge multimodal complementary features by using multiple learnable weight coefficients. Numerous experimental results verify that our DSAFormer outperforms other state-of-the-art methods on multiple public and common datasets.

\ifCLASSOPTIONcaptionsoff
  \newpage
\fi


\bibliographystyle{IEEEtran}
\bibliography{reference}

@misc{YOLOv5,
  author    = {Glenn Jocher},
  title     = {YOLOv5 by Ultralytics},
  year      = {2020},
  note      = {url \url{https://github.com/ultralytics/yolov5}}
}

@misc{YOLOv8,
  author    = {Glenn Jocher and
               Ayush Chaurasia and
               Jing Qiu},
  title     = {Ultralytics YOLO},
  year      = {2023},
  note      = {url \url{https://github.com/ultralytics/ultralytics}}
}

@article{Ren17,
  author       = {Shaoqing Ren and
                  Kaiming He and
                  Ross B. Girshick and
                  Jian Sun},
  title        = {Faster {R-CNN:} Towards Real-Time Object Detection with Region Proposal
                  Networks},
  journal      = {{IEEE} Transactions on Pattern Analysis and Machine Intelligence},
  volume       = {39},
  number       = {6},
  pages        = {1137--1149},
  year         = {2017}
}

@article{Hu25,
  author       = {Yu Hu and
                  Xiaobo Chen and
                  Sheng Wang and
                  Luyang Liu and
                  Hengyang Shi and
                  Lihong Fan and
                  Jing Tian and
                  Jun Liang},
  title        = {Deformable Cross-Attention Transformer for Weakly Aligned {RGB-T}
                  Pedestrian Detection},
  journal      = {{IEEE} Transactions on Multimedia},
  volume       = {27},
  pages        = {4400--4411},
  year         = {2025}
}

@article{Qian25,
  author       = {Jingchen Qian and
                  Baiyou Qiao and
                  Yuekai Zhang and
                  Tongyan Liu and
                  Shuo Wang and
                  Gang Wu and
                  Donghong Han},
  title        = {DACFusion: Dual Asymmetric Cross-Attention guided feature fusion for
                  multispectral object detection},
  journal      = {Neurocomputing},
  volume       = {635},
  pages        = {129913},
  year         = {2025}
}

@article{HuH25,
  author       = {Ke Hu and
                  Yudong He and
                  Yuan Li and
                  Jiayu Zhao and
                  Song Chen and
                  Yi Kang},
  title        = {EI{\({^2}\)}Det: Edge-Guided Illumination-Aware Interactive Learning
                  for Visible-Infrared Object Detection},
  journal      = {{IEEE} Transactions on Circuits and Systems for Video Technology},
  volume       = {35},
  number       = {7},
  pages        = {7101--7115},
  year         = {2025}
}

@article{Fang22,
  author       = {Qingyun Fang and
                  Zhaokui Wang},
  title        = {Cross-modality attentive feature fusion for object detection in multispectral
                  remote sensing imagery},
  journal      = {Pattern Recognition},
  volume       = {130},
  pages        = {108786},
  year         = {2022}
}

@inproceedings{Cao23,
  author       = {Yue Cao and
                  Junchi Bin and
                  Jozsef Hamari and
                  Erik Blasch and
                  Zheng Liu},
  title        = {Multimodal Object Detection by Channel Switching and Spatial Attention},
  booktitle    = {{IEEE} Conference on Computer Vision and Pattern Recognition Workshops},
  pages        = {403--411},
  year         = {2023}
}

@inproceedings{Zhang21,
  author       = {Heng Zhang and
                  {\'{E}}lisa Fromont and
                  S{\'{e}}bastien Lef{\`{e}}vre and
                  Bruno Avignon},
  title        = {Guided Attentive Feature Fusion for Multispectral Pedestrian Detection},
  booktitle    = {{IEEE} Winter Conference on Applications of Computer Vision},
  pages        = {72--80},
  year         = {2021}
}

@article{Xie23,
  author       = {Yumin Xie and
                  Langwen Zhang and
                  Xiaoyuan Yu and
                  Wei Xie},
  title        = {{YOLO-MS:} Multispectral Object Detection via Feature Interaction
                  and Self-Attention Guided Fusion},
  journal      = {{IEEE} Transactions on Cognitive and Developmental Systems},
  volume       = {15},
  number       = {4},
  pages        = {2132--2143},
  year         = {2023}
}

@article{Fang21,
  author       = {Qingyun Fang and
                  Dapeng Han and
                  Zhaokui Wang},
  title        = {Cross-Modality Fusion Transformer for Multispectral Object Detection},
  journal      = {arXiv preprint arXiv:2111.00273},
  year         = {2021}
}

@article{Lee24,
  author       = {Seungik Lee and
                  Jaehyeong Park and
                  Jinsun Park},
  title        = {CrossFormer: Cross-guided attention for multi-modal object detection},
  journal      = {Pattern Recognition Letters},
  volume       = {179},
  pages        = {144--150},
  year         = {2024}
}

@article{Shen24,
  author       = {Jifeng Shen and
                  Yifei Chen and
                  Yue Liu and
                  Xin Zuo and
                  Heng Fan and
                  Wankou Yang},
  title        = {ICAFusion: Iterative cross-attention guided feature fusion for multispectral
                  object detection},
  journal      = {Pattern Recognition},
  volume       = {145},
  pages        = {109913},
  year         = {2024}
}

@inproceedings{Chen22,
  author       = {Yi{-}Ting Chen and
                  Jinghao Shi and
                  Zelin Ye and
                  Christoph Mertz and
                  Deva Ramanan and
                  Shu Kong},
  title        = {Multimodal Object Detection via Probabilistic Ensembling},
  booktitle    = {European Conference on Computer Vision},
  volume       = {13669},
  pages        = {139--158},
  year         = {2022}
}

@inproceedings{Jia21,
  author       = {Xinyu Jia and
                  Chuang Zhu and
                  Minzhen Li and
                  Wenqi Tang and
                  Wenli Zhou},
  title        = {{LLVIP:} {A} Visible-infrared Paired Dataset for Low-light Vision},
  booktitle    = {{IEEE} International Conference on Computer Vision Workshops},
  pages        = {3489--3497},
  year         = {2021}
}

@inproceedings{Liu22,
  author       = {Jinyuan Liu and
                  Xin Fan and
                  Zhanbo Huang and
                  Guanyao Wu and
                  Risheng Liu and
                  Wei Zhong and
                  Zhongxuan Luo},
  title        = {Target-aware Dual Adversarial Learning and a Multi-scenario Multi-Modality
                  Benchmark to Fuse Infrared and Visible for Object Detection},
  booktitle    = {{IEEE} Conference on Computer Vision and Pattern Recognition},
  pages        = {5792--5801},
  year         = {2022}
}

@article{Yang25,
  author       = {Fan Yang and
                  Binbin Liang and
                  Wei Li and
                  Jianwei Zhang},
  title        = {Multidimensional Fusion Network for Multispectral Object Detection},
  journal      = {{IEEE} Transactions on Circuits and Systems for Video Technology},
  volume       = {35},
  number       = {1},
  pages        = {547--560},
  year         = {2025}
}

@article{Dong25,
  author       = {Shuhan Dong and
                  Weiying Xie and
                  Danian Yang and
                  Yunsong Li and
                  Jiaqing Zhang and
                  Jiayuan Tian and
                  Jie Lei},
  title        = {SeaDATE: Remedy Dual-Attention Transformer With Semantic Alignment
                  via Contrast Learning for Multimodal Object Detection},
  journal      = {{IEEE} Transactions on Circuits and Systems for Video Technology},
  volume       = {35},
  number       = {5},
  pages        = {4713--4726},
  year         = {2025}
}

@article{You23,
  author       = {Shuai You and
                  Xuedong Xie and
                  Yujian Feng and
                  Chaojun Mei and
                  Yimu Ji},
  title        = {Multi-Scale Aggregation Transformers for Multispectral Object Detection},
  journal      = {{IEEE} Signal Processing Letters},
  volume       = {30},
  pages        = {1172--1176},
  year         = {2023}
}

@article{ZhangY23,
  author       = {Yan Zhang and
                  Huai Yu and
                  Yujie He and
                  Xinya Wang and
                  Wen Yang},
  title        = {Illumination-Guided {RGBT} Object Detection With Inter- and Intra-Modality
                  Fusion},
  journal      = {{IEEE} Transactions on Instrumentation and Measurement‌},
  volume       = {72},
  pages        = {1--13},
  year         = {2023}
}

@inproceedings{Zhao20,
  author       = {Zixiang Zhao and
                  Shuang Xu and
                  Chunxia Zhang and
                  Junmin Liu and
                  Jiangshe Zhang and
                  Pengfei Li},
  title        = {DIDFuse: Deep Image Decomposition for Infrared and Visible Image Fusion},
  booktitle    = {International Joint Conference on Artificial Intelligence},
  pages        = {970--976},
  year         = {2020}
}

@article{ZhangM21,
  author       = {Hao Zhang and
                  Jiayi Ma},
  title        = {SDNet: {A} Versatile Squeeze-and-Decomposition Network for Real-Time
                  Image Fusion},
  journal      = {International Journal of Computer Vision},
  volume       = {129},
  number       = {10},
  pages        = {2761--2785},
  year         = {2021}
}

@inproceedings{Xu22,
  author       = {Han Xu and
                  Jiayi Ma and
                  Jiteng Yuan and
                  Zhuliang Le and
                  Wei Liu},
  title        = {RFNet: Unsupervised Network for Mutually Reinforcing Multi-modal Image
                  Registration and Fusion},
  booktitle    = {{IEEE} Conference on Computer Vision and Pattern Recognition},
  pages        = {19647--19656},
  year         = {2022}
}

@inproceedings{SunC22,
  author       = {Yiming Sun and
                  Bing Cao and
                  Pengfei Zhu and
                  Qinghua Hu},
  title        = {DetFusion: {A} Detection-driven Infrared and Visible Image Fusion
                  Network},
  booktitle    = {{ACM} International Conference on Multimedia},
  pages        = {4003--4011},
  year         = {2022}
}

@inproceedings{ZhaoB23,
  author       = {Zixiang Zhao and
                  Haowen Bai and
                  Jiangshe Zhang and
                  Yulun Zhang and
                  Shuang Xu and
                  Zudi Lin and
                  Radu Timofte and
                  Luc Van Gool},
  title        = {CDDFuse: Correlation-Driven Dual-Branch Feature Decomposition for
                  Multi-Modality Image Fusion},
  booktitle    = {{IEEE} Conference on Computer Vision and Pattern Recognition},
  pages        = {5906--5916},
  year         = {2023}
}

@inproceedings{Li23,
  author       = {Jiawei Li and
                  Jiansheng Chen and
                  Jinyuan Liu and
                  Huimin Ma},
  title        = {Learning a Graph Neural Network with Cross Modality Interaction for
                  Image Fusion},
  booktitle    = {{ACM} International Conference on Multimedia},
  pages        = {4471--4479},
  year         = {2023}
}

@article{TangD22,
  author       = {Linfeng Tang and
                  Yuxin Deng and
                  Yong Ma and
                  Jun Huang and
                  Jiayi Ma},
  title        = {SuperFusion: {A} Versatile Image Registration and Fusion Network with
                  Semantic Awareness},
  journal      = {{IEEE/CAA} Journal of Automatica Sinica},
  volume       = {9},
  number       = {12},
  pages        = {2121--2137},
  year         = {2022}
}

@inproceedings{Zhang20,
  author       = {Heng Zhang and
                  {\'{E}}lisa Fromont and
                  S{\'{e}}bastien Lef{\`{e}}vre and
                  Bruno Avignon},
  title        = {Multispectral Fusion for Object Detection with Cyclic Fuse-and-Refine
                  Blocks},
  booktitle    = {{IEEE} International Conference on Image Processing},
  pages        = {276--280},
  year         = {2020}
}

@article{Liang23,
  author       = {Mingjian Liang and
                  Junjie Hu and
                  Chenyu Bao and
                  Hua Feng and
                  Fuqin Deng and
                  Tin Lun Lam},
  title        = {Explicit Attention-Enhanced Fusion for RGB-Thermal Perception Tasks},
  journal      = {{IEEE} Robotics and Automation Letters},
  volume       = {8},
  number       = {7},
  pages        = {4060--4067},
  year         = {2023}
}

@article{Zeng24,
  author       = {Yuqiao Zeng and
                  Tengfei Liang and
                  Yi Jin and
                  Yidong Li},
  title        = {MMI-Det: Exploring Multi-Modal Integration for Visible and Infrared
                  Object Detection},
  journal      = {{IEEE} Transactions on Circuits and Systems for Video Technology},
  volume       = {34},
  number       = {11},
  pages        = {11198--11213},
  year         = {2024}
}

@inproceedings{Wang24,
  author       = {Ao Wang and
                  Hui Chen and
                  Lihao Liu and
                  Kai Chen and
                  Zijia Lin and
                  Jungong Han and
                  Guiguang Ding},
  title        = {YOLOv10: Real-Time End-to-End Object Detection},
  booktitle    = {Advances in Neural Information Processing Systems},
  year         = {2024}
}

@article{Wu23,
  author       = {Yuanfeng Wu and
                  Xinran Guan and
                  Boya Zhao and
                  Li Ni and
                  Min Huang},
  title        = {Vehicle Detection Based on Adaptive Multimodal Feature Fusion and
                  Cross-Modal Vehicle Index Using {RGB-T} Images},
  journal      = {{IEEE} Journal of Selected Topics in Applied Earth Observations and Remote Sensing},
  volume       = {16},
  pages        = {8166--8177},
  year         = {2023}
}

@article{WangW24,
  author       = {Huiying Wang and
                  Chunping Wang and
                  Qiang Fu and
                  Dongdong Zhang and
                  Renke Kou and
                  Ying Yu and
                  Jian Song},
  title        = {Cross-Modal Oriented Object Detection of {UAV} Aerial Images Based
                  on Image Feature},
  journal      = {{IEEE} Transactions on Geoscience and Remote Sensing},
  volume       = {62},
  pages        = {1--21},
  year         = {2024}
}

@article{Zhang25,
  author       = {Lu Zhang and
                  Zhiyong Liu and
                  Xiangyu Zhu and
                  Zhan Song and
                  Xu Yang and
                  Zhen Lei and
                  Hong Qiao},
  title        = {Weakly Aligned Feature Fusion for Multimodal Object Detection},
  journal      = {{IEEE} Transactions on Neural Networks and Learning Systems},
  volume       = {36},
  number       = {3},
  pages        = {4145--4159},
  year         = {2025}
}

@inproceedings{Girshick14,
  author       = {Ross B. Girshick and
                  Jeff Donahue and
                  Trevor Darrell and
                  Jitendra Malik},
  title        = {Rich Feature Hierarchies for Accurate Object Detection and Semantic
                  Segmentation},
  booktitle    = {{IEEE} Conference on Computer Vision and Pattern Recognition},
  pages        = {580--587},
  year         = {2014}
}

@inproceedings{Girshick15,
  author       = {Ross B. Girshick},
  title        = {Fast {R-CNN}},
  booktitle    = {{IEEE} International Conference on Computer Vision},
  pages        = {1440--1448},
  year         = {2015}
}

@article{He20,
  author       = {Kaiming He and
                  Georgia Gkioxari and
                  Piotr Doll{\'{a}}r and
                  Ross B. Girshick},
  title        = {Mask {R-CNN}},
  journal      = {{IEEE} Transactions on Pattern Analysis and Machine Intelligence},
  volume       = {42},
  number       = {2},
  pages        = {386--397},
  year         = {2020}
}

@inproceedings{Zhou20,
  author       = {Kailai Zhou and
                  Linsen Chen and
                  Xun Cao},
  title        = {Improving Multispectral Pedestrian Detection by Addressing Modality
                  Imbalance Problems},
  booktitle    = {European Conference on Computer Vision},
  volume       = {12363},
  pages        = {787--803},
  year         = {2020}
}

@article{Zhang19,
  author       = {Lu Zhang and
                  Zhiyong Liu and
                  Shifeng Zhang and
                  Xu Yang and
                  Hong Qiao and
                  Kaizhu Huang and
                  Amir Hussain},
  title        = {Cross-modality interactive attention network for multispectral pedestrian
                  detection},
  journal      = {Information Fusion},
  volume       = {50},
  pages        = {20--29},
  year         = {2019}
}

@article{Zhang23,
  author       = {Jiaqing Zhang and
                  Jie Lei and
                  Weiying Xie and
                  Zhenman Fang and
                  Yunsong Li and
                  Qian Du},
  title        = {SuperYOLO: Super Resolution Assisted Object Detection in Multimodal
                  Remote Sensing Imagery},
  journal      = {{IEEE} Transactions on Geoscience and Remote Sensing},
  volume       = {61},
  pages        = {1--15},
  year         = {2023}
}

@article{LiuL22,
  author       = {Tianshan Liu and
                  Kin{-}Man Lam and
                  Rui Zhao and
                  Guoping Qiu},
  title        = {Deep Cross-Modal Representation Learning and Distillation for Illumination-Invariant
                  Pedestrian Detection},
  journal      = {{IEEE} Transactions on Circuits and Systems for Video Technology},
  volume       = {32},
  number       = {1},
  pages        = {315--329},
  year         = {2022}
}

@article{Yan23,
  author       = {Chaoqi Yan and
                  Hong Zhang and
                  Xuliang Li and
                  Yifan Yang and
                  Ding Yuan},
  title        = {Cross-modality complementary information fusion for multispectral
                  pedestrian detection},
  journal      = {Neural Computing and Applications},
  volume       = {35},
  number       = {14},
  pages        = {10361--10386},
  year         = {2023}
}

@inproceedings{Vaswani17,
  author       = {Ashish Vaswani and
                  Noam Shazeer and
                  Niki Parmar and
                  Jakob Uszkoreit and
                  Llion Jones and
                  Aidan N. Gomez and
                  Lukasz Kaiser and
                  Illia Polosukhin},
  title        = {Attention is All you Need},
  booktitle    = {Advances in Neural Information Processing Systems},
  pages        = {5998--6008},
  year         = {2017}
}

@inproceedings{Dosovitskiy21,
  author       = {Alexey Dosovitskiy and
                  Lucas Beyer and
                  Alexander Kolesnikov and
                  Dirk Weissenborn and
                  Xiaohua Zhai and
                  Thomas Unterthiner and
                  Mostafa Dehghani and
                  Matthias Minderer and
                  Georg Heigold and
                  Sylvain Gelly and
                  Jakob Uszkoreit and
                  Neil Houlsby},
  title        = {An Image is Worth 16x16 Words: Transformers for Image Recognition
                  at Scale},
  booktitle    = {International Conference on Learning Representations},
  year         = {2021}
}

@inproceedings{Liu21,
  author       = {Ze Liu and
                  Yutong Lin and
                  Yue Cao and
                  Han Hu and
                  Yixuan Wei and
                  Zheng Zhang and
                  Stephen Lin and
                  Baining Guo},
  title        = {Swin Transformer: Hierarchical Vision Transformer using Shifted Windows},
  booktitle    = {{IEEE} International Conference on Computer Vision},
  pages        = {9992--10002},
  year         = {2021}
}

@inproceedings{LiuH22,
  author       = {Ze Liu and
                  Han Hu and
                  Yutong Lin and
                  Zhuliang Yao and
                  Zhenda Xie and
                  Yixuan Wei and
                  Jia Ning and
                  Yue Cao and
                  Zheng Zhang and
                  Li Dong and
                  Furu Wei and
                  Baining Guo},
  title        = {Swin Transformer {V2:} Scaling Up Capacity and Resolution},
  booktitle    = {{IEEE} Conference on Computer Vision and Pattern Recognition},
  pages        = {11999--12009},
  year         = {2022}
}

@article{Zhu23,
  author       = {Yaohui Zhu and
                  Xiaoyu Sun and
                  Miao Wang and
                  Hua Huang},
  title        = {Multi-Modal Feature Pyramid Transformer for RGB-Infrared Object Detection},
  journal      = {{IEEE} Transactions on Intelligent Transportation Systems},
  volume       = {24},
  number       = {9},
  pages        = {9984--9995},
  year         = {2023}
}

@inproceedings{Chen23,
  author       = {Jierun Chen and
                  Shiu{-}Hong Kao and
                  Hao He and
                  Weipeng Zhuo and
                  Song Wen and
                  Chul{-}Ho Lee and
                  S.{-}H. Gary Chan},
  title        = {Run, Don't Walk: Chasing Higher {FLOPS} for Faster Neural Networks},
  booktitle    = {{IEEE} Conference on Computer Vision and Pattern Recognition},
  pages        = {12021--12031},
  year         = {2023}
}

@article{Howard17,
  author       = {Andrew G. Howard and
                  Menglong Zhu and
                  Bo Chen and
                  Dmitry Kalenichenko and
                  Weijun Wang and
                  Tobias Weyand and
                  Marco Andreetto and
                  Hartwig Adam},
  title        = {MobileNets: Efficient Convolutional Neural Networks for Mobile Vision
                  Applications},
  journal      = {arXiv preprint arXiv:1704.04861},
  year         = {2017}
}

@article{Boer05,
  author       = {Pieter{-}Tjerk de Boer and
                  Dirk P. Kroese and
                  Shie Mannor and
                  Reuven Y. Rubinstein},
  title        = {A Tutorial on the Cross-Entropy Method},
  journal      = {Annals Of Operations Research},
  volume       = {134},
  number       = {1},
  pages        = {19--67},
  year         = {2005}
}

@inproceedings{Rezatofighi19,
  author       = {Hamid Rezatofighi and
                  Nathan Tsoi and
                  JunYoung Gwak and
                  Amir Sadeghian and
                  Ian D. Reid and
                  Silvio Savarese},
  title        = {Generalized Intersection Over Union: {A} Metric and a Loss for Bounding
                  Box Regression},
  booktitle    = {{IEEE} Conference on Computer Vision and Pattern Recognition},
  pages        = {658--666},
  year         = {2019}
}

@article{DongZ25,
  author       = {Wenhao Dong and
                  Haodong Zhu and
                  Shaohui Lin and
                  Xiaoyan Luo and
                  Yunhang Shen and
                  Guodong Guo and
                  Baochang Zhang},
  title        = {Fusion-Mamba for Cross-Modality Object Detection},
  journal      = {{IEEE} Transactions on Multimedia},
  volume       = {27},
  pages        = {7392--7406},
  year         = {2025}
}

@article{Shen26,
  author       = {Jifeng Shen and
                  Haibo Zhan and
                  Shaohua Dong and
                  Xin Zuo and
                  Wankou Yang and
                  Haibin Ling},
  title        = {Multispectral state-space feature fusion: Bridging shared and cross-parametric
                  interactions for object detection},
  journal      = {Information Fusion},
  volume       = {127},
  pages        = {103895},
  year         = {2026}
}

@article{ZhangL24,
  author       = {Ruiheng Zhang and
                  Lu Li and
                  Qi Zhang and
                  Jin Zhang and
                  Lixin Xu and
                  Baomin Zhang and
                  Binglu Wang},
  title        = {Differential Feature Awareness Network Within Antagonistic Learning
                  for Infrared-Visible Object Detection},
  journal      = {{IEEE} Transactions on Circuits and Systems for Video Technology},
  volume       = {34},
  number       = {8},
  pages        = {6735--6748},
  year         = {2024}
}

@inproceedings{ChenL23,
  author       = {Xuanyao Chen and
                  Zhijian Liu and
                  Haotian Tang and
                  Li Yi and
                  Hang Zhao and
                  Song Han},
  title        = {SparseViT: Revisiting Activation Sparsity for Efficient High-Resolution
                  Vision Transformer},
  booktitle    = {{IEEE/CVF} Conference on Computer Vision and Pattern Recognition},
  pages        = {2061--2070},
  year         = {2023}
}

@inproceedings{Zhu21,
  author       = {Xizhou Zhu and
                  Weijie Su and
                  Lewei Lu and
                  Bin Li and
                  Xiaogang Wang and
                  Jifeng Dai},
  title        = {Deformable {DETR:} Deformable Transformers for End-to-End Object Detection},
  booktitle    = {International Conference on Learning Representations},
  year         = {2021}
}

@inproceedings{Rao21,
  author       = {Yongming Rao and
                  Wenliang Zhao and
                  Benlin Liu and
                  Jiwen Lu and
                  Jie Zhou and
                  Cho{-}Jui Hsieh},
  title        = {DynamicViT: Efficient Vision Transformers with Dynamic Token Sparsification},
  booktitle    = {Advances in Neural Information Processing Systems},
  pages        = {13937--13949},
  year         = {2021}
}

@article{Zhao26,
  author       = {Tianyi Zhao and
                  Maoxun Yuan and
                  Feng Jiang and
                  Nan Wang and
                  Xingxing Wei},
  title        = {Removal Then Selection: {A} Coarse-to-Fine Fusion Perspective for
                  RGB-Infrared Object Detection},
  journal      = {{IEEE} Transactions on Intelligent Transportation Systems},
  volume       = {27},
  number       = {2},
  pages        = {2504--2519},
  year         = {2026}
}

@article{Fan26,
  author       = {Fan Yang and
                  Wei Li and
                  Lei Li and
                  Min Yang and
                  Jianwei Zhang},
  title        = {DWSF-Net: A Dynamic Wavelet-based Spatial-frequency Fusion Network for Multispectral Object Detection},
  journal      = {IEEE Transactions on Multimedia},
  pages={1-13},
  year={2026}
}

@article{Shang25,
  author       = {Xiping Shang and
                  Nannan Li and
                  Dongjin Li and
                  Jianwei Lv and
                  Wei Zhao and
                  Rufei Zhang and
                  Jingyu Xu},
  title        = {CCLDet: {A} Cross-Modality and Cross-Domain Low-Light Detector},
  journal      = {{IEEE} Transactions on Intelligent Transportation Systems},
  volume       = {26},
  number       = {3},
  pages        = {3284--3294},
  year         = {2025}
}

@inproceedings{Yaun25,
  author        = {Maoxun Yuan and
                   Bo Cui and
                   Tianyi Zhao and
                   Jiayi Wang and
                   Shan Fu and
                   Xue Yang and
                   Xingxing Wei},
  title         = {UniRGB-IR: A Unified Framework for Visible-Infrared Semantic Tasks via Adapter Tuning},
  booktitle     = {ACM International Conference on Multimedia},
  pages         = {2409–2418},
  year          = {2025}
}

@inproceedings{Li25,
  author       = {Ke Li and
                  Di Wang and
                  Zhangyuan Hu and
                  Shaofeng Li and
                  Weiping Ni and
                  Lin Zhao and
                  Quan Wang},
  title        = {FD2-Net: Frequency-Driven Feature Decomposition Network for Infrared-Visible
                  Object Detection},
  booktitle    = {{AAAI} Conference on Artificial Intelligence},
  pages        = {4797--4805},
  year         = {2025}
}

@article{Liu25,
  author       = {Yanfeng Liu and
                  Wei Guo and
                  Chaojun Yao and
                  Lefei Zhang},
  title        = {Dual-Perspective Alignment Learning for Multimodal Remote Sensing
                  Object Detection},
  journal      = {{IEEE} Transactions on Geoscience and Remote Sensing},
  volume       = {63},
  pages        = {1--15},
  year         = {2025}
}

\begin{IEEEbiography}[{\includegraphics[width=0.9in,height=1.2in]{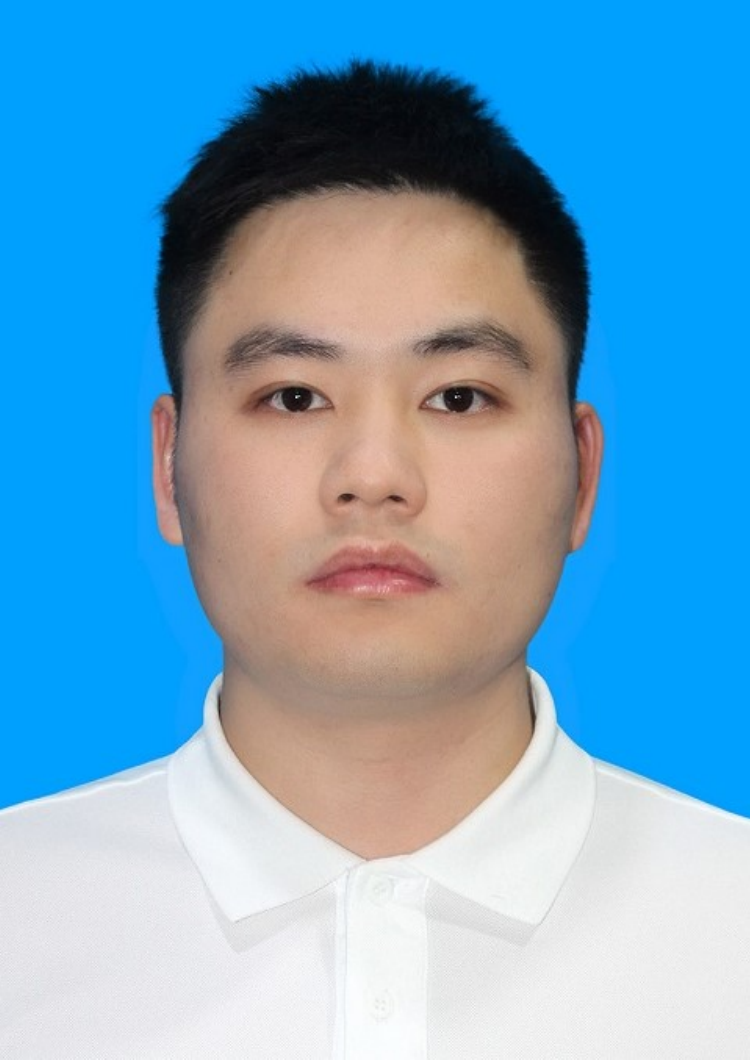}}]{Wencong Wu}
is pursuing the Ph.D. degree at the School of Computer Science, Northwestern Polytechnical University, Xi'an, China. Before, he received the M.S. degree from Yunnan Normal University, Kunming, China, in 2023. His research interests include image restoration and object detection.
\end{IEEEbiography}

\begin{IEEEbiography}[{\includegraphics[width=0.9in,height=1.2in]{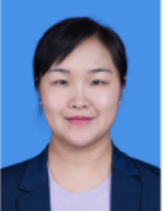}}]{Xiuwei Zhang}
received the B.S., M.S., and Ph.D. degrees from the School of Computer Science, Northwestern Polytechnical University, Xi'an, China, in 2004, 2007, and 2011, respectively.

She is currently a Professor with the School of Computer Science, Northwestern Polytechnical University. Her research interests include remote sensing image processing, multimodel image fusion, image registration, and intelligent forecasting.
\end{IEEEbiography}

\begin{IEEEbiography}[{\includegraphics[width=0.9in,height=1.2in]{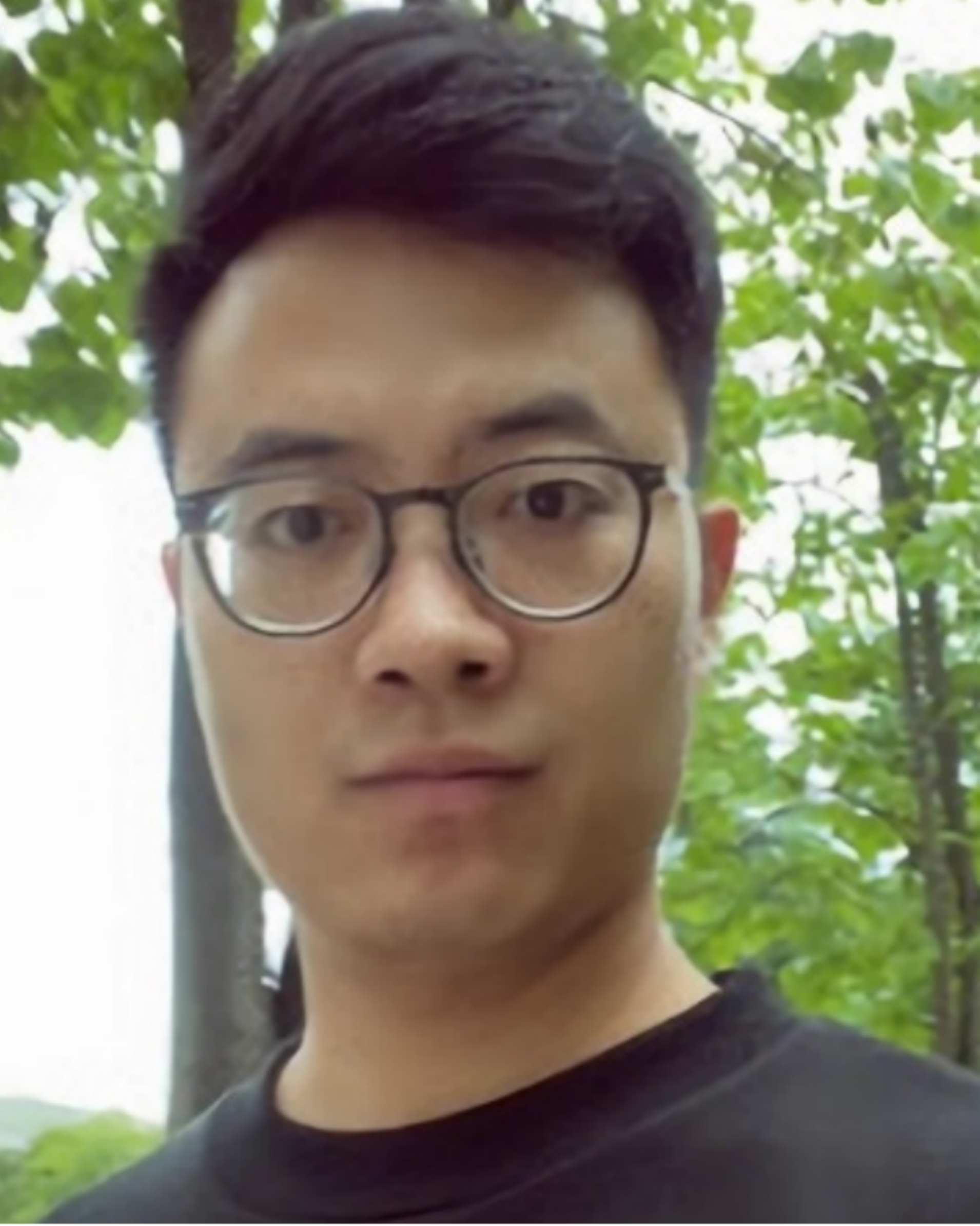}}]{Hanlin Yin}
received the B.S. degree in control science and engineering from Harbin Institute of Technology, Harbin, China, in 2009, and the Ph.D. degree in control science and engineering from Xi'an Jiaotong University, Xi'an, China, in 2016.

He is currently an Assistant Professor with the School of Computer Science, Northwestern Polytechnical University, Xi'an. His research interests include rainfall-runoff modeling, time series prediction, information fusion, and performance evaluation.
\end{IEEEbiography}

\begin{IEEEbiography}[{\includegraphics[width=0.9in,height=1.2in]{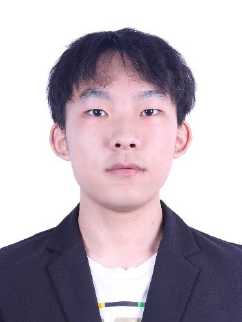}}]{Hongxi Zhang}
received the B.S. degree in Computer Science and Technology from the School of Computer Science, Northwestern Polytechnical University (NWPU), Xi'an, China. He is currently pursuing the M.S. degree with the School of Cybersecurity at NWPU. His research interests include computer vision and multimodal object detection.
\end{IEEEbiography}

\begin{IEEEbiography}[{\includegraphics[width=0.9in,height=1.2in]{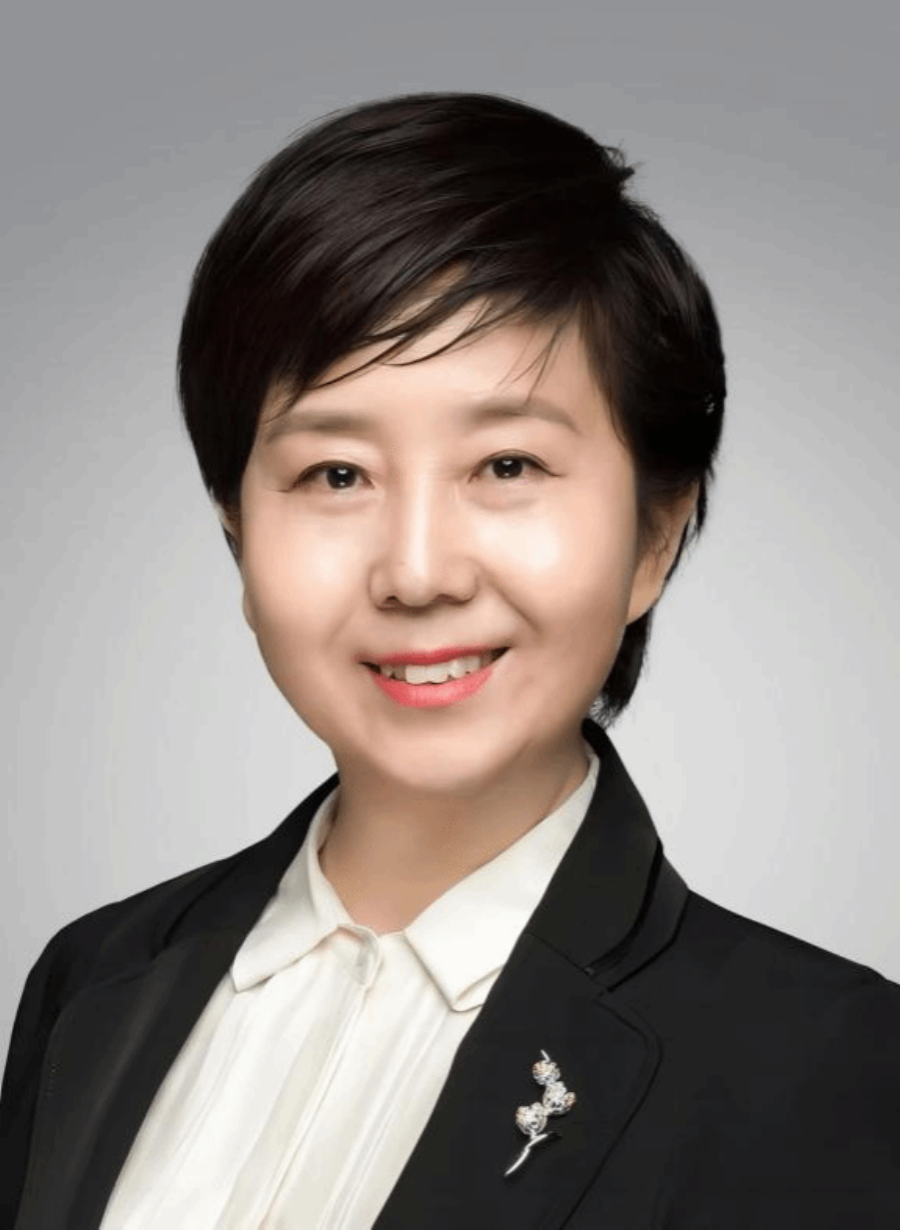}}]{Yanning Zhang} (Fellow, IEEE)
received the B.S. degree from Dalian University of Science and Engineering, Dalian, China, in 1988, and the M.S. and Ph.D. degrees from Northwestern Polytechnical University, Xi'an, China, in 1993 and 1996, respectively.

She is a Professor with the School of Computer Science, Northwestern Polytechnical University. She has published over 200 articles in international journals, conferences, and Chinese key journals. Her research interests include signal and image processing, computer vision, and pattern recognition. Dr. Zhang is also the Organization Chair of the Ninth Asian Conference on Computer Vision (ACCV2009).
\end{IEEEbiography}

\end{document}